\documentclass[graybox]{s-svmult}

\usepackage{mathptmx}       
\usepackage{helvet}         
\usepackage{courier}        
\usepackage{type1cm}        
\usepackage{makeidx}         
\usepackage{graphicx}        
\usepackage{multicol}        
\usepackage[bottom]{footmisc}

\usepackage{booktabs}  
\usepackage{url}
\usepackage{natbib,bibentry}
\usepackage{amsfonts,amssymb,amsmath,bm}
\usepackage{graphics}
\graphicspath{{./Graphics/}}

\usepackage{pgf,pgfplots,pgfplotstable,tikz,tkz-fct}
\usepackage{hyperref}
\hypersetup{
  colorlinks,
  citecolor=blue,
  filecolor=blue,
  linkcolor=blue,
  urlcolor=blue
}

\usepackage{bibunits}

\DeclareMathOperator{\Var}{Var}

\DeclareMathOperator{\MEAN}{E}

\newcommand{\RSS}{{\mathrm{RSS}}}
\newcommand{\R}{\mathfrak{R}}

\newcommand{\RISK}{{\mathrm{R}}}
\newcommand{\Err}{{\mathrm{Err}}}
\newcommand{\TPF}{{\mathrm{TPF}}}
\newcommand{\FNF}{{\mathrm{FNF}}}
\newcommand{\FPF}{{\mathrm{FPF}}}
\newcommand{\AUC}{{\mathrm{AUC}}}
\newcommand{\ROC}{{\mathrm{ROC}}}

\DeclareSymbolFont{letters}{OML}{cmm}{m}{it}

\DeclareMathOperator{\dom}{\mathbf{dom}}

\makeindex             

\usepackage[left=0.65in, right=0.65in, top=0.7in, bottom=0.65in]{geometry}

\begin{document}

\newcommand{\AUTHOR}{Waleed A. Yousef}%
\newcommand{\INSTITUTE}{Waleed A. Yousef\at ECE Dep. ISOT Laboratory, University of Victoria~\email{wyousef@uvic.ca}; and\\CS Dep.,
  HCILAB, Helwan University,~\email{wyousef@fci.helwan.edu}}

\frontmatter

\tableofcontents

\mainmatter


\begin{bibunit}[s-spbasic]
  \title*{Machine Learning Construction: implications to cybersecurity}%
\author{\AUTHOR}%
\institute{\INSTITUTE}%
\maketitle%
\label{ch:MLconstruction}

\abstract{%
  Statistical learning is the process of estimating an unknown probabilistic input-output relationship of a system using a limited
  number of observations. A statistical learning machine (SLM) is the algorithm, function, model, or rule, that learns such a process;
  and machine learning (ML) is the conventional name of this field. ML and its applications are ubiquitous in the modern world.
  Systems such as Automatic target recognition (ATR) in military applications, computer aided diagnosis (CAD) in medical
  imaging, DNA microarrays in genomics, optical character recognition (OCR), speech recognition (SR), spam email filtering, stock
  market prediction, etc., are few examples and applications for ML; diverse fields but one theory. In particular, ML has gained a lot
  of attention in the field of cyberphysical security, especially in the last decade. It is of great importance to this field to design
  detection algorithms that have the capability of learning from security data to be able to hunt threats, achieve better monitoring,
  master the complexity of the threat intelligence feeds, and achieve timely remediation of security incidents. The field of ML can be
  decomposed into two basic subfields: \textit{construction} and \textit{assessment}. We mean by \textit{construction} designing or
  inventing an appropriate algorithm that learns from the input data and achieves a good performance according to some optimality
  criterion. We mean by \textit{assessment} attributing some performance measures to the constructed ML algorithm, along with their
  estimators, to objectively assess this algorithm. \textit{Construction} and \textit{assessment} of a ML algorithm require familiarity
  with different other fields: probability, statistics, matrix theory, optimization, algorithms, and programming, among others. To
  help practitioners, specially those of cyberphysical security, to understand the theoretical foundations of ML, before they delve
  into whole books, we compile the very basics of the first of these two subfields (\textit{construction}) in this chapter. In addition
  to explaining the mathematical foundations of the field, we emphasize the intuitive explanation and concepts.%
}


  \section{Introduction}\label{subsec:introduction}
\subsection{Motivation}\label{sec:motivation-1}
Consider a sample consisting of a number of cases (observations), where each case is composed of a set of inputs and the
corresponding output, all of which will be given to a learning algorithm. Such a sample provides the means for the algorithm to learn
during its so-called \textit{training} (or learning) stage. The goal of this training or learning stage is to understand as much as
possible how the output is related to the inputs in these observations, so that when a new set of inputs is given, in the future, the
algorithm will have some means of predicting the corresponding output. The above terminology has been borrowed from the field of ML.
However, the roots of this problem exists originally in the field of statistical decision theory, where the terminology is somewhat
different. In the latter field, the inputs are called the predictors and the output is called the response. When the output is
quantitative the learning algorithm is called regression; when the output is categorical or ordered categorical the learning algorithm
is called classification. In other communities, the terms \textit{input features} and \textit{output class} are used, respectively. The
learning process can be defined as follows.%
\begin{definition}
  \label{Def_Learning}Learning is the process of estimating an unknown input-output dependency or structure of a system using a limited
  number of observations~\citep{Cherkassky1998LearningFrom}.\quad\qed
\end{definition}

Statistical learning is crucial to many applications. For example, In cyberphysical security, a network activity must be classified as
normal or malicious to avoid any potential threat~\citep{Yousef2021UNAVOIDS}. This is an example of prediction, regardless of whether
it is done by a network analyst or by a ML algorithm. In either case, the prediction is done based on learning from previous network
traffics. The features, i.e., predictors, in this case may be the activity's IP address, number of scanned ports, duration of
connection, etc. The output in this case, i.e., response, is categorical and belongs to the set: $\mathcal{G} =\{normal,\ malicious\}$.
There are so many such examples, including email filtering and spam detection, fraud detection in financial transactions, etc. All of
these examples involve a prediction step based on previous learning.

\bigskip

This chapter reviews some of the regression and classification methods used for predicting a quantitative or categorical response
variable, respectively. In addition, the chapter explains basic concepts related to the performance of these methods. The purpose is
not to present a survey as much as to introduce the field in an approach that combines both mathematics and intuition, and to explain
how the different ingredients relate to each other. We hope this chapter helps practitioners realize the importance of being equipped
with the minimum amount of theory before diving deeply into practice.

\subsection{Notation}\label{sec:notation-1}
Some basic concepts and terminology, necessary for the sequel, must be formally introduced. The world of variables can be categorized
into two categories: deterministic variables and random variables. A deterministic variable takes a definite value; the same value will
be the outcome if the experiment that yielded this value is rerun. On contrary, a random variable is a variable that takes a
non-definite value with a probability value.%
\begin{definition}
  A random variable $X$ is a function from a sample space $S$ into the real numbers $\R$, that associates a real number, $x=X(s)$, with
  each possible outcome $s\in S$.\quad\qed
\end{definition}
Details on the topic can be found in~\cite[Ch. 1]{Casella2002StatisticalInference}. For more rigorous treatment of random variables
based on measure theoretic approach see~\cite{Billingsley1995Probability}. Variables can be categorized as well, based on value, into:
quantitative (or metric), qualitative (or categorical), and ordered categorical. A quantitative variable takes a value on $\R$, and it
can be discrete or continuous. A categorical variable does not necessarily take a numerical value; rather it takes a value from a
finite set. E.g., the set $\mathcal{G}=\{red,\ green,\ blue\}$ is a set of possible qualitative values that can be assigned to a color.
An ordered categorical variable is a categorical variable with relative algebraic relations among the values. E.g., the set
$\mathcal{G}=\{small,\ medium,\ large\}$ includes ordered categorical values.

Variables in a particular process are related to each other in a certain manner. When variables are random the process is said to be
stochastic, i.e., when the inputs of this process have some specified values there is no deterministic value for the output, rather a
probabilistic one. The output in this case is a random variable.

\bigskip

Before delving into mathematical details, it is convenient to introduce some commonly used notation. A random variable---or a random
vector---is referred to by an upper-case letter, e.g., $X$. An instance, case, or observation, of that variable is referred to by a
lower-case letter, e.g., $x$. A collection of $n$ observations for the $p$-dimensional random vector $X$ is collected into an $n\times
p$ matrix and represented by a bold upper-case $\mathbf{X}$. A lower-case bold letter $\mathbf{x}$ is reserved for describing a vector
of any $n$-observations of a variable, even a tuple consisting of non-homogeneous types. The main notation in the sequel will be as
follows: $\mathbf{tr}:\left\{ t_{i}=\left( {x_{i},y_{i}}\right),\ i=1,\ldots,n\right\}$ represents an $n$-case training dataset, i.e.,
one on which the learning mechanism will execute to train, or learn. Every observation $t_{i}$ of this set represents a tuple of the
predictors $x_{i}$ represented in a $p$-dimensional vector, and the corresponding response variable $y_{i}$. All the $n$ observations
$x_{i}$'s may be written in a single $n\times p$ matrix $\mathbf{X}$, while all the observations $y_{i}$ may be written in a vector
$\mathbf{y}$. Some terminologies may arise from diverse scientific communities. To avoid confusion, the word algorithm can be used
exchangeably with function, model, or rule. Using the dataset $\mathbf{tr}$ for learning, training, or fitting, means replacing, or
estimating, the algorithm's unknown parameters with appropriate values, as will be explained throughout the chapter. Therefore, at the
end of this learning process, the final algorithm, function, model, or rule, is called learned, trained, or fitted.

\subsection{Roadmap}\label{sec:roadmap}
The remainder of this chapter is structured as follows. Sec.~\ref{subsec:statistical} introduces the statistical decision theory, which
constitutes the foundation of ML. The chapter explains how the ideal (the best performing) ML algorithm can be constructed, either for
regression or classification, if we know the probability distribution of the data. Sec.~\ref{subsec:parametric} introduces some
important parametric models for both regression and classification, and how they are constructed. Sec.~\ref{subsec:nonparametric}
introduces the nonparametric and smoothing models, and explains the connection to neural network. These three sections will
follow~\cite{Hastie2009ElemStat}, an excellent comprehensive source for regression and classification methods with practical approaches
and illustrative examples. Sec.~\ref{sec:math-optim} introduces mathematical optimization and how it is strongly connected to the
construction of ML algorithms. This section will follow~\cite{Boyd204ConvexOptimization}. Sec.~\ref{sec:performance} discusses, in more
detail, the performance of classification rules. It provides the link between the present and the next chapter.
Sec.~\ref{sec:concl-advice-pract} concludes the chapter and provides a general advice for practitioners.

\section{Statistical Decision Theory}\label{subsec:statistical}
\begin{figure}[t]\centering
  \includegraphics[width = 0.5\textwidth]{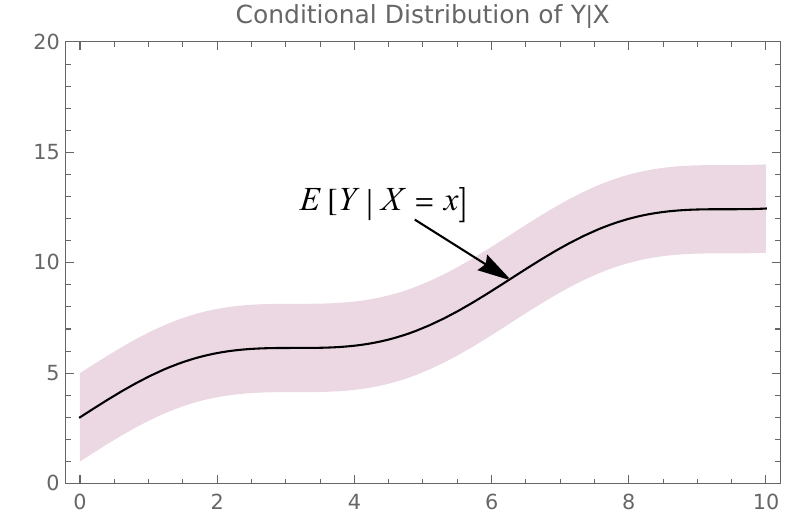}
  \caption{Conditional expectation of a r.v. $Y$, conditional on a r.v. $X$, is the best regression function under the squared-error
    loss.}\label{fig:cond-expect-r.v}
\end{figure}
This section provides an introduction to statistical decision theory, which serves as the foundation of ML. If a random vector $X$ and
a random variable $Y$ have a joint probability density function (PDF) $f_{X,Y}(x,y)$ the problem is defined as follows: how to
predict the variable $Y$ from an observed value for the variable $X$. In this section we assume having a full knowledge of the joint
density $f_{X,Y}$; therefore, there is no learning yet (Definition~\ref{Def_Learning}). The prediction function $\eta(X)$ is required
to have minimum average prediction error. The prediction error should be defined in terms of some loss function $L(Y,\eta(X))$ that
penalizes for any deviation in the predicted value of the response from the correct value. Define the predicted value by:%
\begin{equation}
  \hat{Y}=\eta(X).\label{eq1}
\end{equation}
The risk of this prediction function is defined by the average loss, according to the defined loss function:%
\begin{equation}
  \RISK(\eta)=\MEAN L(Y,\hat{Y}).\label{eq2}%
\end{equation}

\subsection{Regression}\label{sec:regression}
Suppose that the response $Y$ is a quantitative variable. This is the starting point of the statistical branch of regression, where
(\ref{eq1}) is the regression function. A form should be assumed for the loss function. A mathematically convenient and widely used
form is the squared-error loss function:%
\begin{equation}
  L\left(  Y,\eta\left(  X\right)  \right)  =\left(  {Y-\eta\left(  X\right)}\right)^{2}.\label{eq3}%
\end{equation}
In this case (\ref{eq2}) becomes:%
\begin{subequations}
  \begin{align}
    \RISK(\eta)  &  =\int\left(  {Y-\eta(X)}\right)  {^{2}\ dF_{X,Y}(X,Y)}\\
                 &  =\MEAN_{X}  \MEAN_{Y|X}\left[  \left(  {Y-\eta(X)}\right)  {^{2}|X}\right].\label{eq4}%
  \end{align}
\end{subequations}
Hence, (\ref{eq4}) is minimized by minimizing the inner expectation over every possible value for the variable $X$; and the best
regression function is then given by:%
\begin{subequations}
  \begin{align}
    \eta^*(X) & =\arg\min_{\eta(X)}\left[  \MEAN_{Y|X}\left[  \left(  {Y-\eta (X)}\right)  {^{2}|X}\right]  \right]\label{eq5} \\
              & =\MEAN_{Y}\left[Y|X\right]
  \end{align}
\end{subequations}
This means that if the joint distribution for the response and predictor is known the best regression function, in the sense of
minimizing the risk, is the expectation of the response conditional on the predictor (\figurename~\ref{fig:cond-expect-r.v}). In that
case the risk of regression in (\ref{eq4}) will be:%
\begin{equation}
  \RISK_{\min}(\eta)=\RISK(\eta^*)=\MEAN_{X} {\Var}\left[ {Y|X}\right]. \label{eq:3}
\end{equation}

\subsection{Classification}\label{sec:classification}
Recalling (\ref{eq2}), and supposing that the response is a qualitative (or categorical) variable, give rise to the classification
problem. Now the loss function cannot be the squared-error loss function defined in (\ref{eq3}), because this has no meaning for
categorical variables. Because $Y$ may take now a qualitative value from a set of size $K$ (Sec.~\ref{subsec:introduction}), the loss
function can be defined by the matrix%
\begin{equation}
  L(Y,\eta\left(  X\right)  )=\left(  \left(  c_{ij}\right)  \right),\quad 1<i,j<K,\label{eq7}%
\end{equation}
where the non-negative element $c_{ij}$ is the cost, the penalty, or the price, paid for classifying an observation as $y_{j}$ when it
belongs to $y_{i}$. Under this assumption, the risk defined by (\ref{eq2}) can be rewritten for the categorical variables to be:%
\begin{subequations}
  \begin{align}
    \RISK(\eta)  &  =\MEAN_{X}\MEAN_{Y|X}L\left(  {Y,\eta}\left(  {X}\right) \right)\\
                 &  =\MEAN_{X}  \sum\limits_{i=1}^{K}{c_{ij}\Pr}\left[  {Y=y_{i}|X}\right],\label{eq8}%
  \end{align}
\end{subequations}
where $\Pr\left[ Y|X\right] $ is the probability mass function for $Y$ conditional on $X$. Then, the conditional risk for the decision
$y_{j}$,%
\begin{equation}
  \RISK(j,\eta)=\sum\limits_{i=1}^{K}{c_{ij}\Pr}\left[  {Y=y_{i}|X}\right],\label{eq9}
\end{equation}
is the expected loss when classifying an observation as belonging to $y_{j}$, where the expectation is taken over all the possible
values of the response. Again, (\ref{eq8}) can be minimized by minimizing the inner expectation to give:%
\begin{equation}
  \eta^*(X)=\arg\min_{j}\left[  {\sum\limits_{i=1}^{K}{c_{ij}\Pr}}\left[{{Y=y_{i}|X}}\right]  \right].\label{eq10}%
\end{equation}
Expressing the conditional probability of the response in terms of Bayes law, and substituting in (\ref{eq10}) gives:%
\begin{equation}
  \eta^*(X)=\arg\min_{j}\left[ \sum\limits_{i=1}^{K}{c_{ij}f_{X}}\left(  {X|Y=y_{i}}\right)  {\Pr}\left[  {y_{i}}\right]\right].\label{eq11}%
\end{equation}
The probability $\Pr\left[ y_{i}\right] $ is the prior probability for $y_{i}$, while $\Pr\left[ y_{i}|X\right] $ is the posterior
probability, i.e., the probability that the observed case belongs to $y_{i}$, given the value of $X$. This is what is called Bayes
classification, Bayes decision rule, or alternatively, the Bayes classifier.

\bigskip

Some special cases here may be of interest. The first case is when equal costs are assigned to all misclassifications and there is no
cost for correct classification, i.e., $c_{11}=c_{22}=0$ and $c_{12}=c_{21}=1$, which is called the 0-1 cost, or loss function. This
reduces (\ref{eq10}) to:%
\begin{subequations}
  \begin{align}
    \eta^*(X)  &  =\arg\min_{j}\left[  {1-\Pr}[  {Y=y_{j}|X} ]  \right] \label{eq12}\\
               &  =\arg\max_{j}\left[\Pr[Y=y_{j}|X]\right].
  \end{align}
\end{subequations}
The rule thus is to classify the observed case to the class having maximum posterior probability, which is very intuitive.

Another special case of great interest is binary classification, i.e., the case of $K=2$. In this case (\ref{eq10}) reduces to:%
\begin{equation}
  \frac{\Pr\left[  y_{1}|X\right]  }{\Pr\left[  y_{2}|X\right]  }\underset {y_{2}}{\overset{y_{1}}{\gtrless}}\frac{\left(  {c_{22}-c_{21}}\right)}{\left(  {c_{11}-c_{12}}\right)  }.\label{eq13}%
\end{equation}
Alternatively, this can be expressed as:%
\begin{equation}
  \frac{f_{X}(X=x|y_{1})}{f_{X}(X=x|y_{2})}\underset{y_{2}}{\overset{y_{1}}{\gtrless}}\frac{\Pr\left[  y_{2}\right]  \left(  {c_{22}-c_{21}}\right)}{\Pr\left[  y_{1}\right]  \left(  {c_{11}-c_{12}}\right)}.\label{eq14}%
\end{equation}
The decision taken in (\ref{eq10}) has the minimum risk, which can be calculated by substituting back in (\ref{eq8}) to give:%
\begin{equation}
  \RISK_{\min}(\eta)=\sum\limits_{i=1}^{K}{\int_{X}{c_{ij}\Pr}}\left[  {{y_{i}}}\right]  {{dF_{X}(X|y_{i})}},\label{eq15}%
\end{equation}
where $j=\eta(X)$, which is the class decision prediction.

\begin{figure}[t]\centering
  \includegraphics[width = 0.55\textwidth]{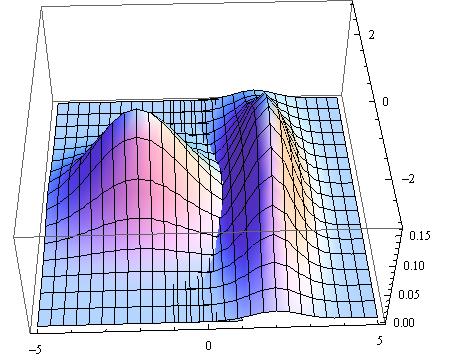}\includegraphics[width = 0.45\textwidth]{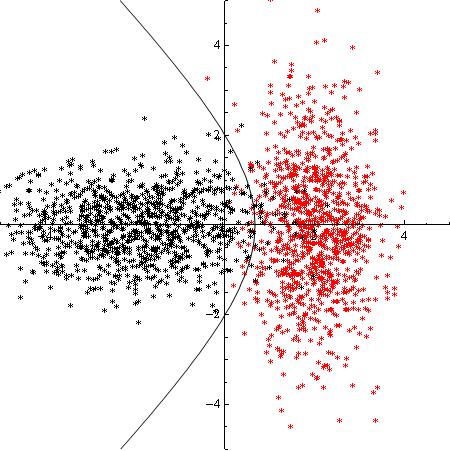}
  \caption{The best decision surface of a binary classification problem with binormal features: the two PDFs, with their intersection
    that shows the best decision surface (left); and how the decision surface looks in the 2D feature space, along with observations
    drawn from the two classes (right).}\label{fig:best-decis-surf}
\end{figure}

For the case where $K=2$ and $c_{ii}=0,\ i=1,2$, Eq.~\eqref{eq15} reduces further to:%
\begin{equation}
  \RISK_{\min}(\eta)=c_{12}\Pr\left[  y_{1}\right]  \int\limits_{R_{2}} {dF_{X}(X|y_{1})}+c_{21}\Pr\left[  y_{2}\right]  \int\limits_{R_{1}} {dF_{X}(X|y_{2})},\label{eq16}%
\end{equation}
where each of $R_{1}$ and $R_{2}$ is the predictor hyperspace over which the optimum decision (\ref{eq13}) predicts as class 1 or class
2, respectively. Later, the response variable Y may be referred to $\Omega$ in case of classification; and to follow the notation of
Sec.~\ref{subsec:introduction}, the response of an observation is assigned a value $\omega_{i},\,i=1,\ldots,K$, to express a certain
class.%
\begin{example}\figurename~\ref{fig:best-decis-surf} illustrates an example of a binary classification problem, where each class has a
  two dimensional predictor, with a binormal distribution, with two different mean vectors $\mu_1,\ \mu_2$, and two different
  covariance matrices $\Sigma_1,\ \Sigma_2$. The best decision surface appears as the intersection of the two PDFs (left). The
  observations sampled from these two classes, along with this best decision surface, are drawn in the 2D space of the predictors
  (right). It is interesting, and may be counter-intuitive for some practitioners, to know that although the two distributions are
  normally distributed, the likelihood ratio~\eqref{eq14} is not necessarily normally
  distributed~\citep{Yousef2020PrudenceWhenAssumingNormality}. For an early development of the theory of binary classification under
  the multinormal assumption of the class distribution,~\cite{Fukunaga1990Introduction} is an indispensable resource.\quad\qed
\end{example}

\subsection{Where Is Learning?}\label{sec:where-learning}
To recap, this section emphasized the fact that there is no distinction between regression and classification from the conceptual point
of view. Each minimizes the risk of predicting the response variable for an observation, i.e., a sample case with known predictor(s).
If the joint PDF for the response and predictors is known, it is just a matter of direct substitution in the above results, which
produces the best regression or classification function that minimizes the risk. If the joint distribution is known but its parameters
are not known, e.g., multinormal distribution with unknown mean vector and covariance matrix, a learning process in this case is
nothing but estimating those parameters from the dataset $\mathbf{tr}$ by well known methods of statistical inference. However, if the
joint distribution is unknown, this gives rise to two different branches of prediction: (1) parametric regression (or classification),
where the regression or classification function is modeled and a training sample is used to build that model, (2) and nonparametric
regression (or classification), where no particular parametric model is assumed. Subsequent sections in this chapter briefly review
some of these techniques, and explain the interesting connections among them.

\section{Parametric Regression and Classification}\label{subsec:parametric}
The prediction method introduced in Sec.~\ref{subsec:statistical} assumes, as indicated, that the joint PDF of the response and the
predictor is known. If such knowledge does not exist all the methods revolve around modeling the regression function (\ref{eq1}) in the
case of regression or the posterior probabilities in (\ref{eq10}) in the case of classification.

\subsection{Linear Models (LM)}\label{subsubsec:linear}
In LM theory, it is assumed that $Y$ is in the form:%
\begin{subequations}
  \begin{align}
    Y  &  =\MEAN Y  +e\label{EqYinLM}\\
       &  =\alpha+{X}^{\prime}\beta+e,\label{EqLM}%
  \end{align}
\end{subequations}
where the randomness of $Y$ comes only from $e$, the conditional expectation of $Y$ is linear in the predictors $X$, and the random
error component $e$ has a zero mean and a constant variance with $X$. The regression function (\ref{eq1}) is then written as:%
\begin{equation}
  \eta(X)=\alpha+{X}^{\prime}\beta.\label{eq17}%
\end{equation}
More generally, still a LM, it can be rewritten as:%
\begin{subequations}\label{eq18EqXnew}
  \begin{align}
    \eta(X)  &  =X_{new}^{\prime}{\beta,}\label{eq18}\\
    X_{new}^{\prime}  &  =\left(  f_{1}\left(  X\right)  ,\ldots,f_{d}\left(X\right)  \right),\label{EqXnew}%
  \end{align}
\end{subequations}
where the predictor $X$ is replaced by a new $d$-dimensional vector, $X_{new}$, whose elements are scalar functions of the original
random vector $X$. The intercept $\alpha$ in (\ref{eq17}) may be absorbed in terms of (\ref{eq18}) by setting $f_{1}\left( X\right)
=1$. Eq. (\ref{eq18}) can be seen as equivalent to (\ref{eq17}), where $X$ has been transformed to $X_{new}$, which became the new
predictor, on which $Y$ will be regressed.

Now $\beta$ must be estimated, and this point estimation is done for some observed values of the predictor; this is merely the learning
process of the LM. Writing the equations for $n$ observed values gives:%
\begin{equation}
  \mathbf{y}=\mathbf{X}\beta+\mathbf{e}.\label{eq21}%
\end{equation}
Eq. (\ref{eq21}) can be solved for $\beta$ to give the least sum-of-squares for the components of error vector $\mathbf{e}$, which is
quite known as the least-squares (LS) problem (Sec.~\ref{sec:math-optim}). Said differently, it can be solved to minimize the residual
sum-of-squares (RSS) between the predicted and the true response:%
\begin{subequations}\label{eq:1}
  \begin{align}
    \RSS &= \mathbf{e}'\mathbf{e}\\
         &= (\mathbf{y}-\mathbf{X}\beta)'(\mathbf{y}-\mathbf{X}\beta)\\
         &= \sum_i(y_i - x_i'\beta)^2,
  \end{align}
\end{subequations}
to give:%
\begin{equation}
  \widehat{\beta}=\left(  \mathbf{X'X}\right)^{-1}\mathbf{X'y}.\label{eq22}%
\end{equation}
Then the prediction $\widehat{Y}$ of $Y$ is done by estimating its expectation, which is given by:%
\begin{equation}
  \widehat{Y} = \widehat{\eta}(X) = \widehat{\MEAN Y} = X\mathbf{^{\prime}}\widehat{\beta}.
\end{equation}
For short notation we always write $\widehat{Y}$ instead of $\widehat{\MEAN\left[ Y\right] }$. The rational behind minimizing the RSS
is that $\RSS/n$ is a good estimate of the mean squared error (MSE), or the expected squared-loss $\MEAN (Y - X'\beta)^2$. In addition,
the latter is differentiable, which leads to the closed-form solution~\eqref{eq22}.

\bigskip

Nothing up to this point involves statistical inference. This is just fitting a mathematical model using the squared-error loss
function. Statistical inference starts when considering the random error vector $\mathbf{e}$ and the effect of that on the confidence
interval for $\hat{\beta}$, and the confidence in predicted values of the response for particular predictor variable, or any other
needed inference. All of these important questions are answered by the theory of LMs.~\cite{Bowerman1990LinearStatistical} is a very
good reference for an applied approach to LMs, without any mathematical proofs. For a theoretical approach and derivations, the reader
is referred to~\cite{Christensen2002PlaneAnswers},~\cite{Graybill1976TheoryAppLinModel}, and~\cite{Rencher2000LinearModels}.

It is remarkable that if the joint distribution of the response and the predictor is multinormal, the LM assumption (\ref{EqLM}) is an
exact expression of the random variable $Y$. This result arises from the fact that the conditional expectation of the multinormal
distribution is linear in the conditional variable. That is, by assuming the joint PDF is multinormal with mean vector $\mu$ and
covariance matrix $\Sigma$, and given by:%
\begin{gather}
  \left( {%
      \begin{array}[c]{l}%
        Y\\
        X
      \end{array}
    }\right)  \sim N\left(  {\mu,\Sigma}\right),\quad%
  \mu=\left( {%
      \begin{array}[c]{l}%
        \mu_{Y}\\
        \mu_{X}%
      \end{array}
    }\right),\quad\Sigma =\begin{pmatrix} \Sigma_{11} & \Sigma_{12}\\ \Sigma_{21} & \Sigma_{22}\end{pmatrix},
\end{gather}
then the conditional expectation of $Y$ on $X$ is given by:%
\begin{equation}
  \MEAN\left[  {\left.  Y\right\vert X=x}\right]  =\mu_{Y}+\Sigma_{12}\Sigma _{22}^{-1}(x-\mu_{X}).\label{eq24}%
\end{equation}
For more details on the multinormal properties see~\cite{Anderson2003AnIntroduction}.

\bigskip

In the case of classification, the classes are categorical variables but a dummy variable can be used as coding for the class labels.
Then a linear regression is carried out for this dummy variable on the predictors. A drawback of this approach is what is called class
masking, i.e., if more than two classes are used, one or more can be masked by others and they may not be assigned to any of the
observations in prediction. For a clear example of masking see~\cite[Sec. 4.2]{Hastie2009ElemStat}.

\subsection{Generalized Linear Models (GLM)}\label{subsubsec:generalized}
In a LM, the response variable is directly related to the regression function by a linear expression of the form (\ref{EqLM}). In many
cases a model can be improved by indirectly relating the response to the predictor through a LM---some times it is necessary, as well,
for the classification problem, as will be shown. This is done through a transformation or a \textit{link} function $g$, by assuming:%
\begin{equation}
  g(\MEAN Y  )=X\mathbf{^{\prime}}\beta.\label{EqYinGLM}%
\end{equation}
Now it is the transformed expectation that is modeled linearly. Hence, LMs are merely a special case of the GLM when the link function
is the identity function $g(\MEAN Y )=\MEAN Y $.

A very useful link function is the \textit{logit} function defined by:%
\begin{equation}
  g(\mu)=\log\frac{\mu}{1-\mu},~~~0<\mu<1.\label{EqLogit}%
\end{equation}
Through this function the regression function is modeled in terms of the predictor as:%
\begin{equation}
  \MEAN\left[  Y\right]  =\frac{\exp(X^{\prime}\beta)}{1+\exp(X^{\prime}\beta)},\label{eq27}%
\end{equation}
which is known as logistic regression (LR). Eq. (\ref{eq27}) implies a constraint on the response $Y$, i.e., it must satisfy
$0<\MEAN\left[ Y\right] <1$, a feature that makes LR an ideal approach for modeling the posterior probabilities in (\ref{eq10}) for the
classification problem. Eq. (\ref{EqLogit}) models the two-class problem, i.e., binary classification, by considering the new responses
$Y_{1}$ and $Y_{2}$ to be defined in terms of the old responses $\omega_{1}\,$and $\omega_{2}$, the classes, as:%
\begin{subequations}
  \begin{align}
    Y_{1}  &  =\Pr\left[  \omega_{1}|X\right]  ,\label{eq28}\\
    Y_{2}  &  =\Pr\left[  \omega_{2}|X\right]  =1-\Pr\left[  \omega_{1}|X\right].
  \end{align}
\end{subequations}
The general case of the $K$-class problem can be modeled using $K-1$ equations, because of the constraint $\sum\nolimits_{k}{\Pr}\left[
  {\omega_{k}|X}\right] =1$, as:%
\begin{equation}
  \log\frac{\Pr\left[  \omega_{k}|X=x\right]  }{\Pr\left[  \omega_{K}|X=x\right]  }=x'\beta_{k},\quad k=1,\ldots,K-1.\label{eq29}%
\end{equation}
Alternatively, (\ref{eq29}) can be rewritten as:%
\begin{align}
  \Pr\left[  \omega_{k}|X=x\right]   &  =\frac{\exp\left(  x'\beta_{k}\right)  }{1+\sum\limits_{k'=1}^{K-1}{\exp\left(  x'\beta_{k'}\right)  }},\quad 1\leq k\leq K-1,\label{eq30}\\
  \Pr\left[  \omega_{K}|X=x\right]   &  =\frac{1}{1+\sum\limits_{k'=1}^{K-1}{\exp\left(  x'\beta_{k'}\right)  }}.
\end{align}
The question now is how to estimate $\beta_{k}$ $\forall$ $k$. The multinomial distribution for modeling observations is appropriate
here. For illustration, consider the case of binary classification; the log-likelihood for the $n$-observations can then be written
as:%
\begin{subequations}
  \begin{align}
    l(\beta)  &  =\sum\limits_{i=1}^{n}\left[ y_i\log\Pr[\omega_1 | X{_{i},\beta}]  + (1-y_{i})\log (1-\Pr[\omega_{1} | X_{i},\beta] )\right]\label{eq31}\\
              &  =\sum\limits_{i=1}^n\left[ y_ix'_i\beta-\log(1+e^{x'_i\beta}) \right].
  \end{align}
\end{subequations}
To maximize this likelihood, the first derivative is set to zero to obtain:%
\begin{equation}
  \frac{\partial l(\beta)}{\partial\beta} = \sum\limits_{i=1}^n x_i\left( y_i-\frac{e^{x'_i\beta}}{1+e^{x'_i\beta}}\right)\overset{set}{=}0.\label{eq32}%
\end{equation}
This is a set of $p$, or $d$, nonlinear equations, because the vector $X$ can be either the original predictor
$(x_{1},\ldots,x_{p})^{\prime}$ or any transformation $(f_{1}(X),\ldots, f_{d}(X))^{\prime}$ as in~(\ref{EqXnew}). These equations can
be solved by iterative numerical methods like the Newton-Raphson algorithm. Finding the optimal values of these parameters is one of
the optimization problems (Sec.~\ref{sec:math-optim}), whose solution exists in many software packages. For more details with numerical
examples see~\cite[Sec. 4.4]{Hastie2009ElemStat} or~\cite[Sec. 12.3]{Casella2002StatisticalInference}.

It can be noted that (\ref{eq31}) is valid under the assumption of the following general distribution:
\begin{equation}
  f(X)=\phi(\theta_{i},\gamma)h(X,\gamma)\exp(\theta_{i}^{\prime}X),
\end{equation}
with probability $p_{i}$, $i=1,2$, $p_{1}+p_{2}=1$, which is the exponential family. So LR is no longer an approximation for the
posterior class probability if the distribution belongs to the exponential family. For insightful comparison between LR and the Bayes
classifier under the multinormal assumption see~\cite{Efron1975TheEfficiencyLogistic}.

\bigskip

It is very important to mention that LR, and all subsequent classification methods, assume equal a priori probabilities. Then the ratio
between the posterior probabilities will be the same as the ratio between the densities that appear in (\ref{eq11}). Hence, the
estimated posterior probabilities from any classification method are used in (\ref{eq11}) as if they are the estimated densities.

\subsection{Nonlinear Models}\label{subsubsec:mylabel1}
The link function in the GLM is modeled linearly in the predictors (\ref{EqYinGLM}). Consequently, the response variable is modeled as
a nonlinear function. In contrast to the LMs described in Sec.~\ref{subsubsec:linear}, in nonlinear models the response can be modeled
nonlinearly right from the beginning, without the need for a link function.

\section{Nonparametric Regression and Classification}\label{subsec:nonparametric}
In contrast to parametric regression, the regression function (\ref{eq1}) is not modeled parametrically; i.e., there is no particular
parametric form to be imposed on the function. Nonparametric regression is a versatile and flexible method of exploring the
relationship of two variables. It may appear that this technique is more efficient than the LMs, but this is not the case. LMs and
nonparametric models can be thought of as two different techniques in the analyst's toolbox. If there is an a priori reason to believe
that the data follow a parametric form, then LMs or parametric regression in general may provide an argument for an optimal choice. If
there is no prior knowledge about the parametric form the data may follow, or no prior information about the physical phenomenon that
generated the data, there may be no choice other than nonparametric regression. There are many nonparametric techniques proposed in the
statistical literature. What was said above, when comparing parametric and nonparametric methods, can also be said when comparing
nonparametric methods to each other. None can be preferred across all situations (Sec.~\ref{sec:concl-advice-pract}).

\subsection{Smoothing Techniques}\label{subsubsec:smoothing}
Smoothing is a tool for summarizing, in a nonparametric way, a trend between a response and a predictor such that the resulting
relationship is less variable than the original response, hence the name smoothing. When the predictor is uni-dimensional, the
smoothing is called scatter-plot smoothing. In this section, some methods used in scatter-plot smoothing are considered. These
smoothing methods do not succeed in higher dimensionality. This is one bad aspect of what is called the curse of dimensionality
(Sec.~\ref{subsec:curse}).

\subsubsection{$K$-Nearest Neighbor (KNN)}\label{para:mylabel1}
The regression function (\ref{eq1}) is estimated in the KNN approach by:%
\begin{gather}
  \eta(x)=\frac{1}{n}\sum\limits_{i=1}^{n}{y_{i}W_{i}(x)},\label{eq34}\\
  W_{i}(x)=\left\{ {%
      \begin{array}[c]{lll}%
        n/K &  & i\in\mathcal{J}_{x}=\left\{  i:x_{i}\in N_{K}(x)\right\} \\
        0 &  & otherwise
      \end{array}
    }\right.,
\end{gather}
where $N_{K}(x)$ is the set consisting of the nearest $K$ points to the point $x$. In words, this technique approximates the
conditional mean, i.e., the regression function that gives minimum risk, by local averaging the response $Y$.

In the case of classification, the posterior probability is estimated by:%
\begin{equation}
  \Pr\left[  \omega_{j}|x\right]  =\frac{1}{n}\sum\limits_{i=1}^n{I_{\omega_{i}=\omega_j}W_i(x)},
\end{equation}
and $I$ is the indicator function defined by:%
\begin{subequations}
  \begin{equation}
    I_{cond}=\left\{  {%
        \begin{array}[c]{l}%
          1\quad cond=True\\
          0\quad cond=False
        \end{array}
      }\right..\label{eq56}%
  \end{equation}
\end{subequations}
That is, replacing the continuous response in (\ref{eq34}) by an indicator function for each class given each observation. So, the
posterior probability is approximated by a frequency of occurrence in a $K$-point neighborhood.

\bigskip

A single-nearest-neighbor method (1-NN) is a special case of the KNN method, where $K=1$. It can be thought of as narrowing the window
$W$ on which regression are carried out. In effect, this makes the regression function or the classifier more complex because it is
trying to estimate the distribution at each point, which results in decreasing the bias and increasing the variance
(Sec.~\ref{subsubsec:variance}).

\subsubsection{Kernel Smoothing}
In this approach, a kernel smoothing function $\kappa$ is assumed. This means that a weighting and convolution (or mathematical
smoothing) is carried out for the points in the neighborhood of the predicted point according to the chosen kernel function. Formally
this is expressed as:%
\begin{equation}
  \eta(x)=\sum\limits_{i=1}^{n}{y_{i}\kappa\left(  \frac{x-x_{i}}{h_{x}}\right)\biggl/\sum\limits_{i^{\prime}=1}^{n}{\kappa}\left(  {\frac {x-x_{i^{\prime}}}{h_{x}}}\right)}\biggr..\label{eq36}
\end{equation}
Choosing the bandwidth $h_{x}$ of the kernel function is not an easy task. Usually, it is done numerically by cross validation (as
explained in the next chapter). It is worth remarking that KNN smoothing is nothing but a kernel smoothing for which the kernel
function is an unsymmetrical flat window spanning the range of the $K$-nearest neighbors of the point $x$. The kernel (\ref{eq36}) is
called Nadaraya-Watson kernel. Historically, and interestingly,~\cite{Parzen1962OnEstimation} first introduced the window method
density function estimation; his work was pioneered later by~\cite{Nadaraya1964EstReg} and~\cite{Watson1964SmoothReg} in regression.

\subsection{Additive Models (AM)}\label{subsubsec:additive}
Recalling (\ref{eq18EqXnew}), and noticing that the function $f_{i}(X)$ is a scalar parametric function of the whole predictor, show
that LMs are parametric AMs. By dropping the parametric assumption and letting each scalar function be a function of just one element
of the predictor, i.e., $X_{i}$, allows defining a new nonparametric regression method, namely AMs, as:
\begin{equation}
  \eta(x)=\alpha+\sum\limits_{i=1}^{p}{f_{i}(X_{i})},\label{eq37}%
\end{equation}
where the predictor is of $p$ dimensions. The response variable itself, $Y$, is modeled as in (\ref{EqYinLM}) by assuming zero mean and
constant variance for the random component $e$. Then, $f_{i}(X_{i})$ is fit by any smoothing method defined in
Sec.~\ref{subsubsec:smoothing}. Every function $f_{i}(X_{i})$ fits the value of the response minus the contribution of the other $p-1$
functions from the previous iteration. This is called the back-fitting algorithm~\cite[Sec. 4.3]{Hastie1990Generalized}

\subsection{Generalized Additive Models (GAM)}\label{subsubsec:mylabel2}
GAMs can be developed in a way analogous to how GLMs were developed above, i.e., by working with a transformation of the response
variable, hence the name generalized additive models. Eq. (\ref{eq37}) describes the regression function as an AM; alternatively it can
be described through another link function:%
\begin{equation}
  g\left(  \eta(x)\right)  =\alpha+\sum\limits_{i=1}^{p}{f_{i}(X_{i})}.\label{eq38}%
\end{equation}

Again, if a \textit{logit} function is used the model can be used for classification exactly as was done in the case of GLMs. Rewriting
the score equations (\ref{eq32}) for the GAM, using the posterior probabilities as the response variable, produces the nonparametric
classification method using the GAM. Details of fitting the model can be found in~\cite[Sec. 4.5 and Ch. 6]{Hastie1990Generalized}.

\subsection{Projection Pursuit Regression (PPR)}\label{subsubsec:projection}
PPR, introduced by~\cite{Friedman1981ProjectionPursuit}, is a direct attack on the dimensionality problem, since it considers the
regression function as a summation of terms, each of which is a function of a projection of the whole predictor onto a direction
(specified by some unit vector). Formally it is expressed as:%
\begin{equation}
  \eta(x)=\sum\limits_{i=1}{g_{i}({\alpha}_{i}^{\prime}x)}.\label{eq39}
\end{equation}
The function $g_{i}$, for every selection of the direction $\alpha_{i}$, is to be fit by a smoother in the new single variable
$\alpha_{i}^{\prime}x$. It should be noted that (\ref{eq39}) assumes that the function $g_{i}(\alpha _{i}^{\prime}X)$, named the
\textit{ridge function}, is constant along any direction perpendicular to $\alpha_{i}$. Fitting the model is done by iteratively
finding the best directions $\alpha_{i}$'s that minimize(s) the RSS, hence the name pursuit. Details of fitting the model and finding
the best projection directions can be found in~\cite{Friedman1981ProjectionPursuit} and~\cite{Hastie2009ElemStat}.

In (\ref{eq39}), by deliberately setting each unit vector $\alpha_{i}$ to have zero components except $\alpha_{ii}=1$, reduces the PPR
to AM. Moreover, and interestingly as well, introducing the \textit{logit} link function to the regression function $\eta(x)$ in
(\ref{eq39}) suits the classification problem exactly as was done in the GAM. This turns out to be exactly the same as the
single-hidden-layer NN, as will be presented in the next section.

\subsection{Neural Networks (NN)}\label{subsubsec:neural}
\begin{figure}[t]\centering
  \tikzset{%
    every neuron/.style={circle, draw, minimum size=0.5cm }, neuron missing/.style={ draw=none, scale=2, text height=0.333cm, execute
      at begin node=\color{black}$\vdots$ }, }
  \begin{tikzpicture}[x=.9cm, y=.9cm, >=stealth]
    \foreach \m/\l [count=\y] in {1,2,3,missing,4} \node [every neuron/.try, neuron \m/.try] (input-\m) at (0,2.5-\y) {};

    \foreach \m [count=\y] in {1,missing,2} \node [every neuron/.try, neuron \m/.try ] (hidden-\m) at (2,2-\y*1.25) {};

    \foreach \m [count=\y] in {1,missing,2} \node [every neuron/.try, neuron \m/.try ] (output-\m) at (4,1.5-\y) {};

    \foreach \l [count=\i] in {1,2,3,p} \draw [<-] (input-\i) -- ++(-1,0) node [above, midway] {$X_\l$};

    \foreach \l [count=\i] in {1,M} \node [above] at (hidden-\i.north) {$Z_\l$};

    \foreach \l [count=\i] in {1,K} \draw [->] (output-\i) -- ++(1,0) node [above, midway] {$Y_\l$};

    \foreach \i in {1,...,4} \foreach \j in {1,...,2} \draw [->] (input-\i) -- (hidden-\j);

    \foreach \i in {1,...,2} \foreach \j in {1,...,2} \draw [->] (hidden-\i) -- (output-\j);

    \foreach \l [count=\x from 0] in {input\ , hidden\ , output\ } \node [align=center, above] at (\x*2,2) {\l layer};
  \end{tikzpicture}\hfill\includegraphics[width = 0.5\textwidth]{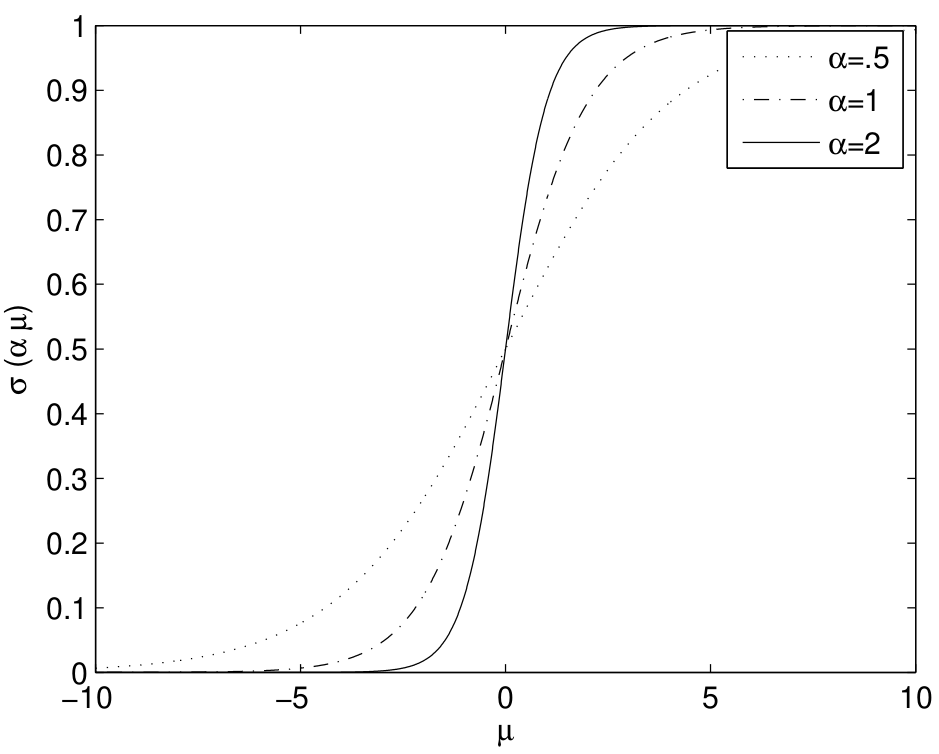}%
  \caption{A single-hidden-layer NN. The architecture that reflects Eq.~\eqref{eq:NN} (left), and the sigmoid function with different
    learning rates $a$ (right).}\label{fig1}%
\end{figure}
The field of NN has been evolving, since its start in the engineering community around 1950s, until we reached now the era of deep
neural networks (DNN). A single-hidden-layer NN can be considered as a process for modeling the output in terms of a linear combination
of the inputs. The set of $p$ input features, i.e., the predictor components $X_{1},\ldots,X_{p}$, are weighted linearly to form a new
set of $M$ arguments, $Z_{1},\ldots,Z_{M}$, that go through the sigmoid function $\sigma$, which can have different values of
steepness, or learning rate. \figurename~\ref{fig1} illustrates a single-hidden-layer NN with its architecture (left), and a plot of
its sigmoid function with different learning rates (right). The output of the sigmoid function accounts for a hidden layer consisting
of $M$ intermediate values. Then these $M$ hidden values are in turn weighted linearly to form a new set of $K$ arguments that go
through the final output functions, whose output is the response variables $Y_{1},\ldots,Y_{K}$. This can be expressed mathematically
in the form:%
\begin{subequations}\label{eq:NN}
  \begin{align}
    Z_{m}  &  =\sigma(\alpha_{om}+{\alpha}_{m}^{\prime}X),\quad m=1,\ldots,M,\\
    \sigma(\mu) &=\frac{1}{1+e^{-\mu}},\label{eq41}\\
    Y_{k}  &  =f_{k}\left(  \beta_{0k}+\sum\limits_{m=1}^{M}{\beta_{mk}Z_{m}}\right),\quad k=1,\ldots,K.\label{EqNN}%
  \end{align}
\end{subequations}
Eq. (\ref{EqNN}) shows that if the function $f$ is chosen to be the identity function, i.e., $f(\mu)=\mu$, the NN is simply a special
case of the PPR method defined in (\ref{eq39}), where the sigmoid function has been explicitly imposed on the model rather than being
developed by any smoothing mechanism as in PPR. This is what is done when the output of the network is quantitative. When it is
categorical, i.e., the case of classification, the function $f$ can be simply modeled as:%
\begin{equation}
  f_{k}(\mu_{k})=e^{\mu_{k}}\biggl/\sum\limits_{{k}^{\prime}=1}^{K}{e^{\mu_{{k}^{\prime}}}}\biggr..\label{eq42}%
\end{equation}
In this case each output node models the posterior probability $\Pr\left[ \omega_{k}|X\right] $, which is exactly what is done by the
LR link function defined in (\ref{EqLogit}). Again, the model will be an extension to the GAM as defined at the end of
Sec.~\ref{subsubsec:projection}. Although equations~\eqref{eq:NN} are indeed parametric, we list NN in this section for the strong
connection to the AM, GAM, and PPR that were just explained. Excellent references for the early basics and foundations of NN
are~\cite{Bishop1995NeurNet} and~\cite{Ripley1996PRandNN}. We conclude this section by quoting the following statement
from~\cite{Hastie2009ElemStat}:

\begin{quotation}
  ``There has been a great deal of hype surrounding neural networks, making them seem magical and mysterious. As we make clear in this
  section, they are just nonlinear statistical models, much like the projection pursuit regression model discussed above.''
\end{quotation}


  \section{Optimization}\label{sec:math-optim}
Optimization serves an amazing variety of practical problems: e.g., optimizing power consumption in electrical stations, optimizing
overall budget in project management, and most importantly to us in this chapter optimizing ML algorithms to provide the best
performance. In this section, we will provide a very basic introduction to optimization and its strong connection to the construction
of ML algorithms.

\subsection{Introduction}\label{sec:intr-defin}
The mathematical optimization problem (MOP) is an abstraction of how to make the ``best'' possible choice of some vector $\beta$ under
some \textit{constraints}. These constraints represent a set of trim requirements, or specifications, that limits the possible choices
of this vector. The \textit{objective function} of this problem represents the \textit{cost}, or \textit{loss}, to minimize, or the
\textit{utility} to maximize, for each vector $\beta$, and this what makes that value of $\beta$ the ``best'' possible choice. This is
formalized in the following definition.
\begin{definition}\textit{A mathematical optimization problem} has the form:\label{def:intr-defin}%
  \setlength{\belowdisplayskip}{0pt} \setlength{\abovedisplayskip}{0pt}
  \begin{align*}
    &\underset{\beta}{\textup{minimize}} &&f_0(\beta)\\
    &\textup{subject to:} && f_i(\beta) \leq 0,\quad i=1,\ldots,m,\\
    & && h_i(\beta) = 0, \quad i=1,\ldots,l,
  \end{align*}
  where%
  \begin{align*}
    \beta &=(\beta_1,\ldots,\beta_p)\in \R^p, \tag{\textit{optimization variable}}\\
    f_0&:\ \R^p \mapsto \R, \tag{\textit{objective (cost) function}}\\
    f_i&:\ \R^p \mapsto \R, \tag{\textit{inequality constraints (functions)}}\\
    h_i&:\ \R^p \mapsto \R, \tag{\textit{equality constraints (functions)}}\\
    \mathcal{D}&:\ \bigcap_{i=1}^{m} \dom f_i\ \cap\ \bigcap_{i=1}^l \dom h_i\tag{\textit{domain of constraints: feasible set}}\\
          &= \left\{\beta\ |\ \beta\in \R^p\ \wedge\ f_i(\beta) \leq 0\ \wedge\ h_i(\beta)=0\right\}\\
    \beta^*:&\left\{\beta\ |\ \beta \in \mathcal{D}\ \wedge\ f_0(\beta) \leq f_0(\alpha)\ \forall \alpha \in \mathcal{D}\right\},  \tag{\textit{solution}}
  \end{align*}
  where the solution $\beta^*$ is called the optimizer (or minimizer).\quad\qed
\end{definition}
The problem aims at minimizing a mathematical function, under some constraints. From definition~\ref{def:intr-defin}, it is clear that
minimizing $f_0$ is the same problem as maximizing $-f_0$; the constraints $f_i\leq 0$ are equivalent to $-f_i\geq 0$; the constraints
$f_i \leq 0$ are equivalent to $f_i \leq b_i$, where $b_i$ can be simply absorbed into $f_i$; and, finally, $m=l=0$ is the case of
unconstrained problem with global minimization.%

\begin{figure}[t] \centering%
  \usepgfplotslibrary{fillbetween} \pgfmathdeclarefunction{line}{0}{\pgfmathparse{4*(x-1)+1}}
  \begin{tikzpicture}[scale=0.75]
    \begin{axis}[ axis y line = center, axis x line = bottom, samples = 160, domain = -4:4, xmin = -4, xmax = 4, ymin = 0, ymax = 10,]
      \filldraw[fill=orange!70, draw=white] (500,0) rectangle ++(100,100); \addplot[name path=poly, black, thick, mark=none, ] {x^2};
    \end{axis}
  \end{tikzpicture}
  \caption{An objective function in a single dimension, with a constraint $1 \leq \beta \leq 2$ (the colored region). The minimizer
    $\beta^* = 1$, under this constraint, is different from the global minimizer $\beta^* = 0$.}\label{fig:optimization-example}
\end{figure}
\begin{example}: The is a very basic example of an MOP in a single dimension, with a single constraint:%
  \begin{align*}\label{eq:2}
    &\underset{\beta}{\textup{minimize}} && f_0(\beta) = \beta^2\\
    &\textup{subject to:} && \beta \leq 2,\\
    & && 1 \leq \beta.
  \end{align*}
  It is clear that the minimizer is $\beta^* = 1$; however, the minimizer for the unconstrained problem is $\beta^* = 0$
  (\figurename~\ref{fig:optimization-example}).\quad\qed
\end{example}

\begin{figure}[t]\centering
  \begin{tikzpicture}[scale=0.5]
    \begin{axis}[ 3d box, width=15cm, view={25}{25}, enlargelimits=false, grid=major, domain=-6:6, y domain=-6:6, samples=21,
      xlabel=$\beta_1$, ylabel=$\beta_2$, zlabel={$f_0$} ]

      \addplot3 [y domain = -6:6, surf, opacity=0.5, faceted color = yellow] {(x-1)^2+y-2};

      \addplot3 [y domain = -6:6, contour gnuplot = {levels={-0.25},number=14, labels={false}, draw color = black}, samples = 50]
      {(x-1)^2+y-2};

      \addplot3 [contour gnuplot = {levels={1,-0.25,-2.5}, number=14, labels={false}, draw color=black}, samples=50, z
      filter/.code={\def\pgfmathresult{60}}] {(x-1)^2+y-2};

      \addplot3+ [samples=20, no markers, color = red] (x,1+x,60);

      \addplot3+ [samples=20, no markers, color=red] (x,2-x,60);
    \end{axis}
  \end{tikzpicture}
  \caption{A simple objective function in two dimensions (the colored surface, shown along with its contours drawn in black), with two
    constraints (the red lines). Although the surface has no global minimum, the constrained problem does
    have.}\label{fig:an-object-funct}
\end{figure}
\begin{example}\citep[Ex. 20.1, P. 454]{Chong2001AnIntorductionToOptimization}: This example shows how the MOP may not be
  as simple as finding the derivatives:%
  \begin{align*}
    &\underset{\beta}{\textup{minimize}} && f_0(\beta_1,\beta_2) = (\beta_1-1)^2 + \beta_2 -2\\
    &\textup{subject to:} && \beta_2-\beta_1 = 1,\\
    & &&\beta_1+\beta_2 \leq 2.%
  \end{align*}
  This 2D objective function, along with the constraints, are illustrated in~\figurename~\ref{fig:an-object-funct}. It is obvious that
  the function has no global minimizer ($\partial f_0/\partial \beta_2 = 1 \neq 0$). After setting the constraints, it is quite easy to
  see that $f_0|_{(\beta_2-\beta_1 = 1)} = (\beta_1-1)^2+(\beta_1-1)$ attains a minima at $\beta_1 = 1/2$, and hence, the minimizer is
  $\beta^* = (1/2, 3/2)'$.\quad\qed
\end{example}

\subsection{Connection to Machine Learning}\label{sec:motiv-appl}
As explained earlier in this chapter, all parametric ML algorithms, e.g., LM, LR, SVM, NN, DNN, etc., include parameters that need to
be replaced by numerical values. This is performed with the help of a dataset that is called a training dataset. The following simple
example illustrates the connection between ML and MOP, and relates both to the title of the present chapter.%

\bigskip

\begin{figure}[t]\centering \includegraphics[width=0.35\textwidth]{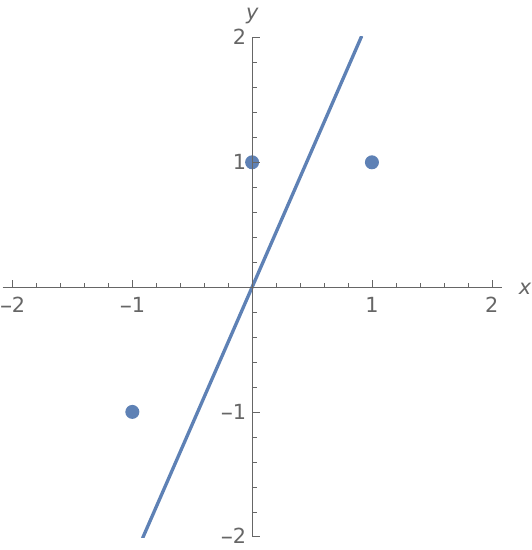}\hfil\includegraphics[width=0.65\textwidth, trim = 0 60
  0 60, clip]{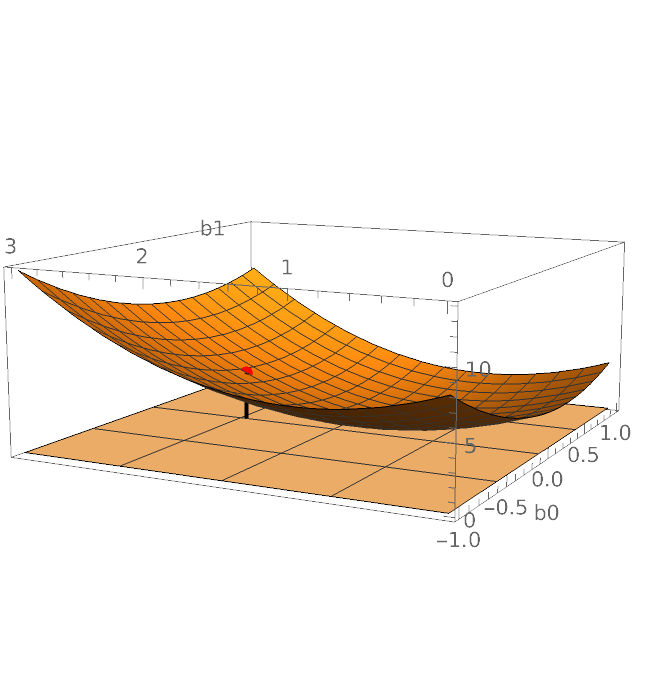}
  \caption{ML and MOP. A training dataset of three observations, for a regression problem with a single feature, to be fitted by a
    linear model having only two parameters $\beta_0$ and $\beta_1$ (left). The RSS of this model is the objective function, of these
    two parameters, to be minimized (right). The red point on the surface is the value (not the minimum yet), at some initial values of
    $\beta_0$ and $\beta_1$ that corresponds to the intercept and slope of the line on the left.}\label{fig:optim-train-datas}
\end{figure}
\begin{example}[Machine Learning: construction]\label{ex:conn-mach-learn-1} Suppose that we have a strong belief that the best
  regression function for a particular problem is the LM $Y = \beta_0 + \beta_1 X$. Then, for a given training dataset
  $\mathbf{tr}:\left\{ t_{i}=\left( {x_{i},y_{i}}\right),\ i=1,\ldots,n\right\}$, we need to minimize the RSS of this
  model on this dataset. This is a typical MOP, which can be formalized as:%
  \begin{align*}
    &\underset{\beta_0,\beta_1}{\textup{minimize}} \sum_{i=1}^n (\beta_0 + \beta_1 x_i - y_i)^2,\quad (x_i, y_i) \in \mathbf{tr}.
  \end{align*}
  \figurename~\ref{fig:optim-train-datas} illustrates a dataset of only three observations ($n=3$), along with a straight line of
  initial values of the parameters $\beta_0,\ \beta_1$, all drawn in the feature space (left). The objective function to be minimized
  (the RSS) is drawn as a function of the two parameters $\beta_0, \beta_1$ (right). Each pair of values of
  $\beta_0,\ \beta_1$ results in a new line (fitted model) in the feature space, and a new point on the surface of the objective function in
  the parameter space. The solution of this MOP is the vector $\mathbf{\beta}^* = (\beta_0,\beta_1)$ that minimizes the objective
  function, which fortunately for linear models has a closed-form solution given earlier in Eq.~\eqref{eq22}.\quad\qed
\end{example}

\bigskip

\begin{figure}[t]\centering%
  \includegraphics[width=0.5\textwidth]{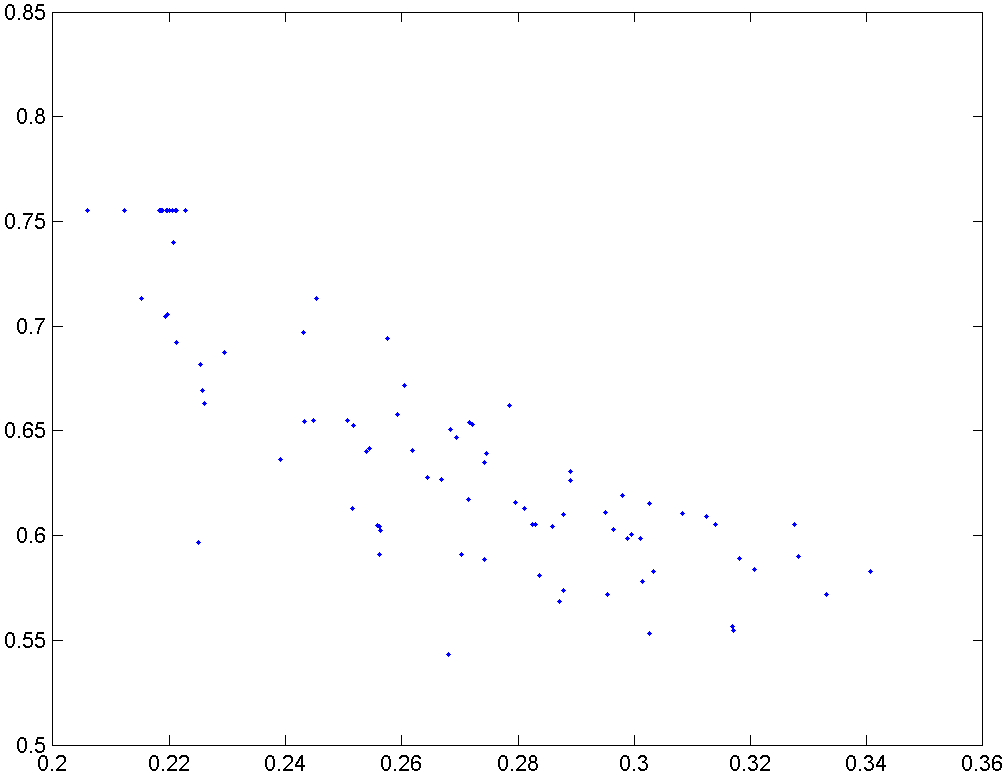}
  \caption{100 pairs of true AUC vs. true MSE. Each pair is obtained from training the same NN, to minimize the RSS, on a new training
    datasets, then testing on a very large testing dataset to mimic the population. Although there is an obvious trend of getting a
    high AUC with low MSE, it is not a guaranteed behaviour for each training dataset.\label{fig:100-pairs-true}}
\end{figure}
Departing from the previous example, in the following few paragraphs we will emphasize important concepts. The example demonstrated the
relationship between: (1) the ML model, along with the training dataset, in the feature space, and (2) the objective function, which
should be minimized, in the parameter space. All other ML models, whether for regression or classification, have parameters that should
be replaced, tuned, or estimated, to optimize (minimize or maximize) some objective function. Ideally, this objective function should
be a good estimator for the same intended performance measure of the model, not for any other performance measure. However,
some mathematical difficulties may preclude this ideal practice, as will be seen next.

In Example~\ref{ex:conn-mach-learn-1}, the objective function to be minimized was the RSS, which minimizes $\RSS/n$, an estimator of
the MSE. However, in some circumstances it is very difficult mathematically to optimize the targeted performance measure. In such
cases, another performance measure is optimized, because of the tractability of its mathematical formalization, hoping that the
solution optimizes, as well, the targeted performance measure. Figure~\ref{fig:100-pairs-true} illustrates this fact for a very simple
four-neuron single-layer NN, trained on a simulated two-class univariate normal dataset. The NN is required to achieve a high AUC (a
performance measure that will be explained in Sec.~\ref{sec:performance}); however, because its estimator is non differentiable, and
therefore is very hard to maximize using conventional mathematical approaches, the NN is trained to minimize the RSS, instead. For
illustration, the NN is trained on 100 different training datasets, and tested after each training on a very large testing dataset to
provide a good estimate of the true AUC and MSE. The figure shows the 100 pairs of values of these two performance measures, with a
general trend of exhibiting a high AUC with a low MSE. However, some instances exhibited a low MSE (good performance) associated with a
low AUC (bad performance).

\bigskip

Another important fact to emphasize is that all of the model's parameters that are estimated during the learning process are functions
of the training dataset. Hence, the following facts hold: these parameters are random variables, the model is a random model, the
objective function is a random function, and the minimum value of this objective function, which is the model optimal performance, is a
random variable (as will be detailed in Sec.~\ref{sec:performance}); all will vary if the training dataset varies.

\bigskip

To recap, example~\ref{ex:conn-mach-learn-1} demonstrated how LM, one of the ML models explained in this chapter, represents an MOP
whose solution fortunately can be found in closed form. Other ML models belong to a class of MOP that is difficult to solve; DNN is
an example. In the next section, we will provide a very short account of the taxonomy of the MOP to show its different types and the
connection of each of these types to ML.

\subsection{Types of MOP}\label{sec:solv-optim-probl}
According to the nature of the objective function and its constraints, the MOP can be classified into one of these nested classes:

\hfil Linear $\subset$ Quadratic $\subset$ Convex $\subset$ Nonlinear.\hfil\\For each of these classes, several questions arise: (1) is
there a closed-form solution? (2) if not, is there a numerical solution? (3) if yes, is it guaranteed? (4) what are the ML models that
belong to this class? In the following subsections, we discuss very briefly each of these classes, and provide some answers to these
questions. It is important to emphasize that although these classes are mathematically nested---in the very strict sense that any
linear is quadratic, any quadratic is convex, and any convex is nonlinear---the solution techniques for each class are quite different
from others. The solution techniques for these classes vary between: closed-form, numerical (e.g., Newton's methods, gradient descent,
etc.), or even \textit{intelligent}-based methods (e.g. genetic algorithms, particle swarm, etc.).

\subsubsection{Linear Programming}\label{sec:linear-programming}
\begin{definition}A linear programming problem is an MOP with an objective and all constraints are
  linear:\label{def:linear-programming}%
  \begin{align*}
    & \underset{\beta}{\textup{minimize}} &  & f_0(\beta)= c'\beta                         \\
    & \textup{subject to:}            &  & a_i'\beta \leq b_i,\quad i=1,\ldots,m, \\
    &                                 &  & h_i'\beta = g_i,\quad i=1,\ldots,l.\quad\qed
  \end{align*}
\end{definition}
\begin{example}[\textit{Chebyshev minimization}] The MOP:%
  \[\underset{\beta}{\textup{minimize}} f_0(\beta) = \max_{i=1,\ldots,n} |y_i - x_i'\beta|,\]
  can be understood in terms of ML terminology as minimizing the maximum possible error, measured in absolute deviance between the true
  response value $y_i$ and the predicted value $x_i'\beta$. This should be contrasted with the least-squares MOP of the LM (as
  illustrated in example~\ref{ex:conn-mach-learn-1} and will be more detailed in Sec.~\ref{sec:least-squar-probl} below) in two
  important aspects: (1) the error is measured in terms of absolute deviance rather than squared difference. (2) the objective
  function here focuses only on the single observation that achieves the maximum error rather than summing over all observations. After
  little manipulations, the problem can be reduced, and found to be equivalent, to the following:%
  \begin{align*}
    &\underset{\beta}{\textup{minimize}} &&t\\
    &\textup{subject to:} && x_i'\beta-t \leq y_i, &&& i=1,\ldots,n\\
    & && -x_i'\beta-t \leq-y_i, &&& i=1,\ldots,n,
  \end{align*}
  which is a typical linear programming problem, per definition~\ref{def:linear-programming}.\quad\qed
\end{example}

\bigskip

In general, there is no closed-form solution to the linear programming problems. However, there exists a set of very robust, reliable,
and computationally effective methods of numerical solutions: e.g., Dantzig's simplex and interior point that can solve problems with
several thousands of variables.

\subsubsection{Least-Squares (LS) Problems}\label{sec:least-squar-probl}
\begin{definition}A LS problem is an MOP with no constraints (i.e., $m=l=0$), and an objective in the
  form:\label{def:least-squar-probl}%
  \[\underset{\beta}{\textup{minimize}} f_0(\beta) = \sum_{i=1}^n (x_i'\beta - y_i)^2 = \|\mathbf{X}_{n \times p}\beta_{p \times 1} - \mathbf{y}_{n \times
      1}\|^2.\quad\qed\]
\end{definition}
\begin{example}[LM] The linear models for regression, discussed in Sec.~\ref{subsubsec:linear} and Example~\ref{ex:conn-mach-learn-1},
  is a typical example for the LS problem, where the solution is given in the closed form by
  $\beta = \left( \mathbf{X'X}\right) ^{-1}\mathbf{X'y}$.\quad\qed
\end{example}

\bigskip

The algorithms for finding the matrix inversion and matrix multiplication in this closed-form solution exist in many scientific
computing software, and this technology is quite mature even for thousands of variables.

\bigskip

There is a more elaborate version of the LS problem that is called weighted LS. This type of problem appears in ML, e.g., when more
emphasis is required on some observations than others:%
\[\underset{\beta}{\textup{minimize}} f_0(\beta) = \sum_{i=1}^n w_i (x_i'\beta - y_i)^2,\]
or when it is required to penalize for using extra parameters to guard against overfitting (Sec.~\ref{subsubsec:variance}), an
approach known as regularization:%
\[\underset{\beta}{\textup{minimize}} f_0(\beta) = \sum_{i=1}^n (x_i'\beta - y_i)^2 + \rho \sum_{j=1}^p \beta_j^2.\]
It is quite easy to show that both problems can be solved as LS problem, per definition~\ref{def:least-squar-probl}.

\subsubsection{Convex Optimization}\label{sec:convex-optimization}
\begin{definition}A convex optimization problem is an MOP with an objective and all constraints are convex:%
  \begin{align*}
    & \underset{\beta}{\textup{minimize}} &  & f_0(\beta)                  \\
    & \textup{subject to:}            &  & f_i(\beta) \leq 0, &&& i=1,\ldots,m, \\
    &                                 &  & h_i(\beta) = 0 ,    &&& i=1,\ldots,l,\\
    & &&f_i(a \alpha + b \beta) \leq a f_i(\alpha) + b f_i(\beta), &&& a + b =1,\quad 0 \leq a,b, \quad 0 \leq i \leq m,\\
    & && h_i(\beta) = c_i'\beta+d_i &&&0 \leq i \leq p.\quad\qed
  \end{align*}
\end{definition}
\begin{example}[Lasso Regression] Similar to the penalized LS problem, lasso regression minimizes the RSS; however, it does so with an
  $L_1$ penalty rather than the $L_2$ of the LS problem. The problem is formalized as:%
  \begin{align*}
    &\underset{\beta}{\textup{minimize}} &&\sum_{i=1}^n (y_i - x_i'\beta)^2\\
    &\textup{subject to:} && \sum_{j=1}^p |\beta_j| \leq t,
  \end{align*}
  which can be shown to be equivalent to the MOP:%
  \[\underset{\beta}{\textup{minimize}} f_0(\beta) = \sum_{i=1}^n (y_i - x_i'\beta)^2 + \rho \sum_{j=1}^p |\beta_j|,\]
  The latter, in contrast to the LS penalization, has no closed-form solution because of the difficulty introduced by
  the non-differentiable term $|\beta_j|$. However, the numerical solution is quite feasible and reliable as all convex optimization
  problems are.\quad\qed
\end{example}

\bigskip

The numerical solution of a convex optimization problem is well established through the methods of \textit{interior point}, although no
closed-form solution exists. Problems with thousands of variables can be solved robustly as in linear programming problems. In
addition, many problems are initially formulated, then with some mathematical manipulation they can be transformed to a solvable convex
problem.

\subsubsection{Nonlinear Optimization}\label{sec:nonl-optim}
\begin{figure}[t]\centering \includegraphics[width=0.75\textwidth]{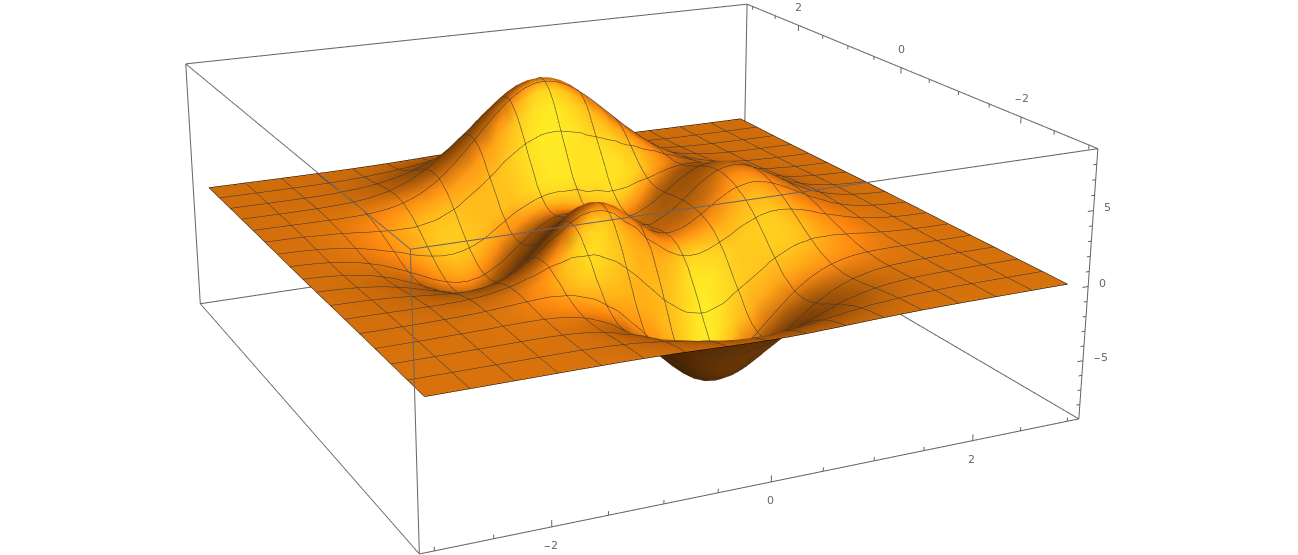}
  \caption{A plot of a 3D function~\citep[Ex. 14.3, P.290]{Chong2001AnIntorductionToOptimization}: $f(\beta_0,\beta_1) = 3
    (1-\beta_0)^2 e^{-\beta_0^2-(\beta_1+1)^2}-10 e^{-\beta_0^2-\beta_1^2}
    \left(-\beta_0^3+\frac{\beta_0}{5}-\beta_1^5\right)-\frac{1}{3} e^{-(\beta_0+1)^2-\beta_1^2}$ that shows several minima, maxima,
    and saddle points.}.\label{fig:3d-plot-function}
\end{figure}
\begin{definition}A nonlinear optimization problem is an MOP with objective and constraint functions are nonlinear\quad\qed
\end{definition}
\begin{example} A nonlinear objective function, just in two dimensions, with several minima, maxima, and saddle points, is illustrated
  in~\figurename~\ref{fig:3d-plot-function}. A NN, or in its more complex form, a DNN with several layers, can have hundreds of millions of
  parameters, not only two as illustrated in the figure!\quad\qed
\end{example}

The nonlinear optimization problems can be very hard to solve, even for simple-looking problems in few parameters (variables). Several
approaches exist for solving the problem; these approaches can be divided into two main categories: numerical methods and
computational intelligence, as briefly explained in the following two paragraphs, respectively.

Finding the minima or the maxima of a function numerically is a well known topic in mathematics and numerical analysis. However, the
challenge of nonlinear optimization, especially for problems like DNN, remains in the computational complexity that grows exponentially
with the dimensions of the objective function, the matter that makes it almost impossible to find a global minimum. Alternatively,
finding a local minimum is a practical compromise, although it does not guarantee converging to the global one. Local minimization
starts at a point in the parameter space (usually is selected randomly, or by other criteria determined by the numerical algorithm)
then the space is navigated, and guided by the multi-dimensional derivatives (with respect to the parameters) of the objective
function. All the well known methods, starting form Newton's method to the most recent approaches used for DNN, e.g., stochastic
gradient descent (SGD), belong to this category. It is obvious that the initial starting point in the parameter space heavily affects
the convergence process and the final solution.

The term computational intelligence was first coined early by~\cite{Bezdek1992onRelationship,Bezdek1994WhatIsCI}:%
\begin{quote}
  ``A system is computationally intelligent when it: deals only with numerical (low-level) data, has a pattern recognition component,
  and does not use knowledge in the AI (Artificial Intelligence) sense; and additionally, when it (begins to) exhibit (i) computational
  adaptivity; (ii) computational fault tolerance; (iii) speed approaching human-like turnaround, and (iv) error rates that approximate
  human performance.''
\end{quote}
Since that time, the term computational intelligence (CI) has been accepted as a generic term to the field that combines NNs, fuzzy
logic, and evolutionary algorithms~\citep{Schwefel2003AdvancesIn,Zimmermann2002AdvCI}. Later, the area of swarm detection was
considered as a peer paradigm to the other three mentioned above~\cite{Engelbrecht2002Computational}.


  \section{Performance}\label{sec:performance}
From what has been early discussed at the beginning of this chapter, there is not any conceptual difference between regression and
classification for the problem of supervised learning. Abstractly, both aim to achieve the minimum risk~\eqref{eq2} under a certain
loss function, for predicting a response, from a particular predictor. Although risk is a very obvious performance measure for
assessing ML algorithms, we will elaborate in this section and show how we can depart and define other important performance measures,
e.g., the individual error components, ROC, and AUC. It is must be noted that what will be defined in this section is the parametric
form (also known as the true performance or the population performance), which can only be calculated if the posterior probabilities
are known. On the contrary, if the posterior probabilities are not known all performance measures can be estimated from a given
dataset, called the testing dataset, using appropriate estimators. If the testing dataset is infinitely large, i.e. testing on the
population, the estimated performance will converge to the true performance. Performance estimation and different estimators are
discussed in the next chapter.

\subsection{Error Components}\label{sec:error-components-1}
\begin{figure}[t]\centering
  \includegraphics[width = 0.5\textwidth]{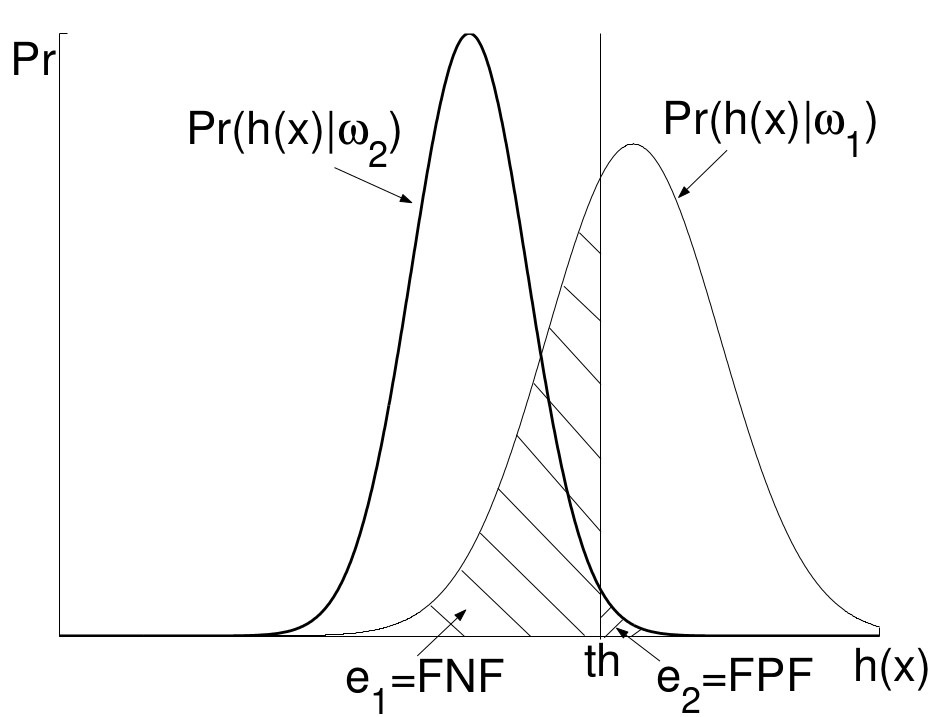}%
  \caption{The probability of LLR conditional on each class. The two components of error are indicated as the FPF and
    FNF.}\label{fig3}%
\end{figure}
We will elaborate on the special case of binary classification, with no cost on correct classification ($c_{ii}=0,\ i=1,2$), which
is of great interest in many applications. In this case, the risk of each classifier is reduced to (\ref{eq16}), which can be rewritten
as:%
\begin{equation}
  \RISK_{\min}=c_{12}P_{1}e_{1}+c_{21}P_{2}e_{2},\label{eq44-01}
\end{equation}
where $e_{1}$ is the probability of classifying a case as belonging to class 2 when it belongs to class 1, and $e_{2}$ is vice versa.

In the feature space, the regions of classification have the dimensionality $p$, and it is very difficult to calculate the error
components from multi-dimensional integration. It is easier to look at (\ref{eq14}) as:%
\begin{subequations}\label{eq:45All}
  \begin{align}
    &  h(x)\underset{\omega_{2}}{\overset{\omega_{_{1}}}{\gtrless}}th,\label{eq45}\\
    h(x)  &  =\log\frac{f_{X}(X=x|\omega_{1})}{f_{X}(X=x|\omega_{2})},\\
    th  &  =\log\frac{\Pr[\omega_{1}] c_{21}}{\Pr[\omega_{2}] c_{12}  },
  \end{align}
\end{subequations}
where the $\log$ is taken just as a convention to simplify the analysis for the case of multinormal distribution (because it has an
exponent); however, it has no other significance. The function $h(X)$ is called the log-likelihood ratio (LLR), which is obviously a
random variable, whose variability comes from the feature vector $X$. The LLR has a PDF conditional on each of the two classes, as
indicated in~\figurename~\ref{fig3}; (it can be easily shown that the two curves in this figure cross at $h(X)=0$, when the threshold is
zero.)

In general, the two error components appearing in (\ref{eq44-01}) can be rewritten, equivalently to their corresponding terms in
Eq.~\eqref{eq16}, using the LLR in~\eqref{eq:45All}, as:%
\begin{subequations}\label{Eqe1e2-01}
  \begin{align}
    e_{1}  &  =\int_{-\infty}^{th}{f_{h}}\left(  {h(x)|\omega_{1}}\right) {dh(x)},\\
    e_{2}  &  =\int_{th}^{\infty}{f_{h}}\left(  {h(x)|\omega_{2}}\right)  {dh(x)}.
  \end{align}
\end{subequations}

\bigskip

Now, it is very important to realize the generality of this error equation and the two messages it conveys. (1) It expresses the two
components of error for any classifier that produces an output, or a score, of $h(x)$ for a predictor $X = x$, even if it is not the
best (Bayes') classifier. The only exception then would be that the score $h(X)$ is no longer the LLR that produces the minimum risk.
(2) Whether $h(X)$ is the score of the Bayes' classifier or not, Eq.~\eqref{Eqe1e2-01} says that at each threshold value $th$ there is
a pair of two components of error. Over the continuum of threshold values there is a continuum of these pairs, which define a new
curve. This curve is called the ROC curve, a device that is much more rich for assessing classification rules than a single pair of
errors, as will be explained next.

\subsection{Receiver Operating Characteristic (ROC) Curve}\label{sec:r}
\begin{figure}[t]\centering
  \includegraphics[width=0.5\textwidth]{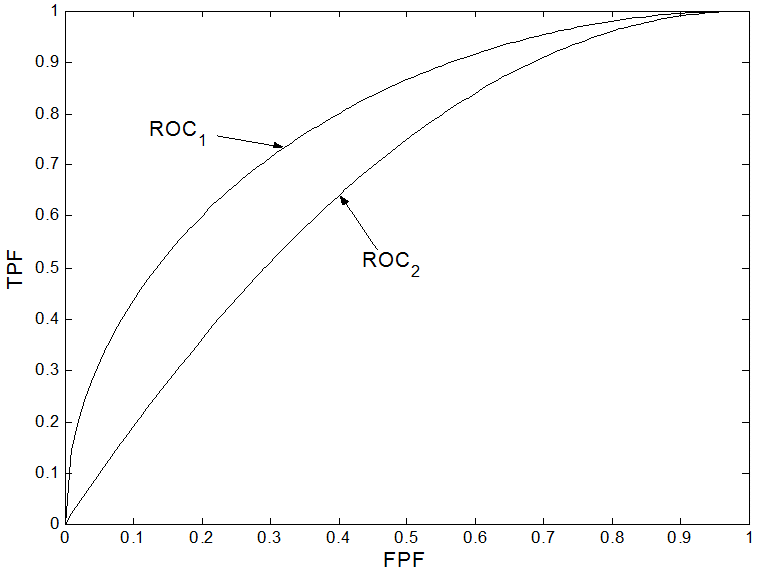}\includegraphics[width=0.5\textwidth]{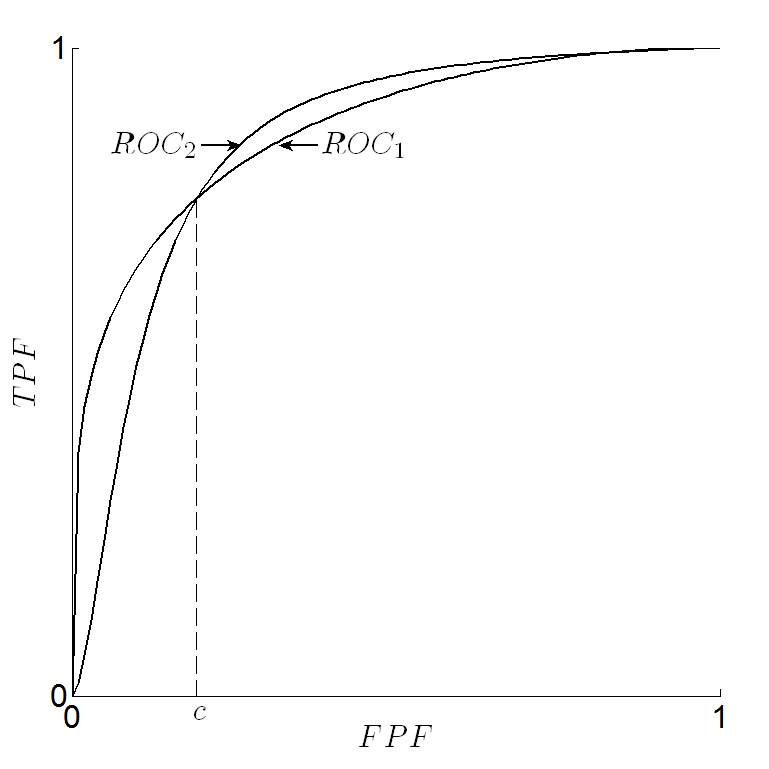}
  \caption{ROC curves for two different competing classifiers. Left: ROC$_{1}$ is better than ROC$_{2}$, since for any error component
    value, the other component of classifier 1 is less than that of classifier 2. Right: ROC$_{1}$ is better than ROC$_{2}$ only in the
    range of FPF that is lower than the value $c$.}\label{fig4-01}
\end{figure}
Now, assume the classifier is trained under the condition of equal prevalence and cost, i.e., the threshold is zero. In other
environments there will be different a priori probabilities yielding to different threshold values. The error is not a sufficient
metric now, since it is a function of a single fixed threshold. A more general way to assess a classifier is provided by the ROC curve.
This is a plot for the two components of error, $e_{1}$ and $e_{2}$, under different threshold values. It is conventional in many
applications to refer to $e_{1}$ as the False Negative Fraction (FNF), and $e_{2}$ as the False Positive Fraction (FPF). This is
because cases from the abnormal class typically are assigned higher classifier's scores than cases from the normal class, hence the
names ``positive'' and ``negative''. For example, a network activity belonging to the abnormal class (the class of anomalous
activities) whose classifier's score is less than the chosen threshold will be called ``negative''. This is obviously a false negative
decision; hence the name FNF. The situation is reversed for the other error component.

Because the classification problem now can be seen, more generally, in terms of the classifier's output score rather than the hard
binary decision, it is apparent that each of the two error components is an integral over a univariate PDF. Therefore, the resulting
ROC is a monotonically non-decreasing function. A convention in many fields is to plot the true positive fraction (TPF), which is given
by $\TPF=1-\FNF$, vs. the $\FPF$. In that case, the farther apart the two distributions $f_h(h|\omega_i),\ i=1,2$ of the score function
$h(X)$ from each other, the higher the ROC curve and the larger the area under the curve (AUC).~\figurename~\ref{fig4-01} (left) shows
ROC curves for two different competing classifiers. The first classifier performs better because it has a lower value of $e_{2}$ at
each value of $e_{1}$. Thus, the first classifier unambiguously separates the two classes better than the second one. Therefore, the
AUC for the first classifier is larger than that for the second one. The AUC can be thought of as one summary performance measure for
the ROC curve. Formally, the AUC is given by:%
\begin{equation}
  \AUC=\int_{0}^{1}{\TPF~d(\FPF)}.\label{eq47-01}%
\end{equation}
And it can be shown that it is also given by:%
\begin{equation}
  \AUC=\Pr\bigl[ h(x)|\omega_{2} < h(x)|\omega_{1}\bigr],\label{eq:PopulationManWhit-01}
\end{equation}
which expresses how the classifier scores for class $\omega_1$ are stochastically larger than those of class $\omega_2$, and hence more
capable of the classification task.

If two ROC curves cross (\figurename~\ref{fig4-01}, right), this means each classifier is better than the other only for a certain
range of the threshold setting, and vice versa. In that case, some other performance measure can be used, such as the partial area
under the ROC curve in a specified region~\citep{Yousef2013PAUC}.

\subsection{The True Performance Is A Random Variable!}\label{sec:role-train-test}
\begin{figure}[t]\centering
  \includegraphics[width = 0.5\textwidth]{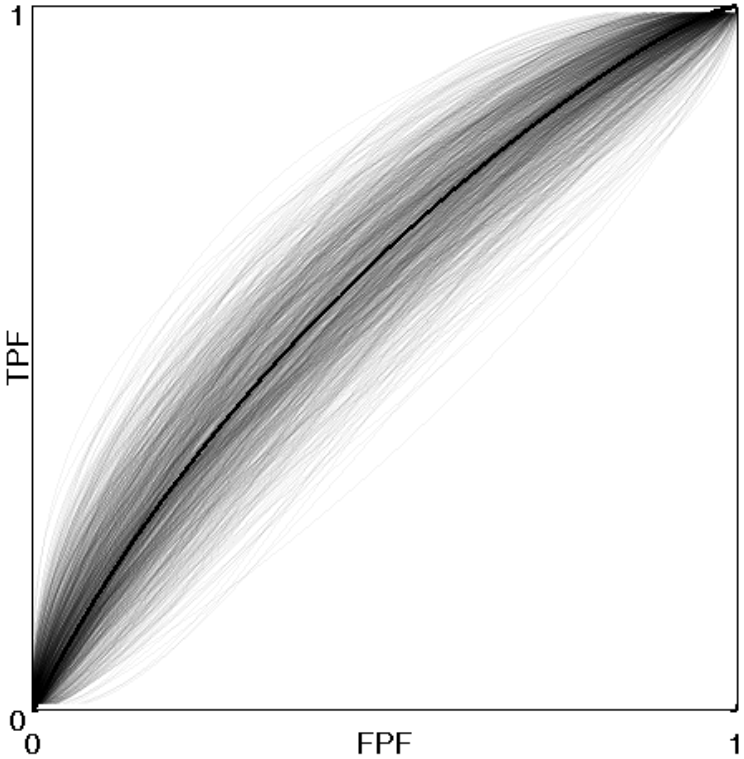}
  \caption{A population of ROC curves of a classifier trained on several training datasets; for each training dataset an ROC is built.
    The mean ROC is shown in bold.}\label{fig:popul-roc-curv}
\end{figure}
As was explained in Sec.~\ref{sec:motiv-appl}, regardless of whether the ML task is regression or classification, the model, its
parameters, and its performance, all are random variables, where the randomness comes from the training dataset. In addition, this
variation depends on the complexity of the model, and its capacity to learn, relative to the training dataset size.

For instance, the output scoring function $h(X)$ of a particular classifier is indeed $h_{\mathbf{tr}}(X)$, which is subscripted to show
the dependence on the training dataset; hence, the PDFs $f_{h_{\mathbf{tr}}}(h|\omega_i)$ and $\ROC_{\mathbf{tr}}$, all should be
subscripted as well. Therefore, there is a population of ROC curves that corresponds to the population of training datasets
(\figurename~\ref{fig:popul-roc-curv}). For more elaboration, consider the AUC as the performance measure of interest. Then, the
fundamental quantities of interest are the following:%
\begin{enumerate}
\item $\AUC_{{\rm {\bf tr}}} $: the true performance of the classifier, conditional on a particular training dataset ${\rm {\bf tr}}$ of
  a specified size $n$ but over the population of testing datasets (as if we trained on ${\rm {\bf tr}}$ then tested on infinite number
  of observations),

\item $\MEAN_{{\rm {\bf tr}}} \AUC_{{\rm {\bf tr}}} $: the expectation of the true performance over the population of training datasets
  of the same size $n$, and

\item $\Var_{\rm {\bf tr}}\AUC_{{\rm {\bf tr}}}$: the variance of the true performance over the population of training datasets of the
  same size $n$. This variance expresses how the classifier is sensitive to retraining, e.g. in the case of obtaining a new training
  dataset.
\end{enumerate}
Any other performance measure, e.g., each of the error components, the risk, etc., is a r.v. as well, should be similarly subscripted
$\mathbf{tr}$, and has a mean and a variance as explained above. For more elaboration, we explain this crucial concept in
Sec.~\ref{subsubsec:variance}, in a more mathematical detail, for the case of regression, because it is more obvious and easier to
explain.

\subsection{Bias-Variance Decomposition}\label{subsubsec:variance}
\begin{figure}[t]\centering
  \includegraphics[width = 0.5\textwidth]{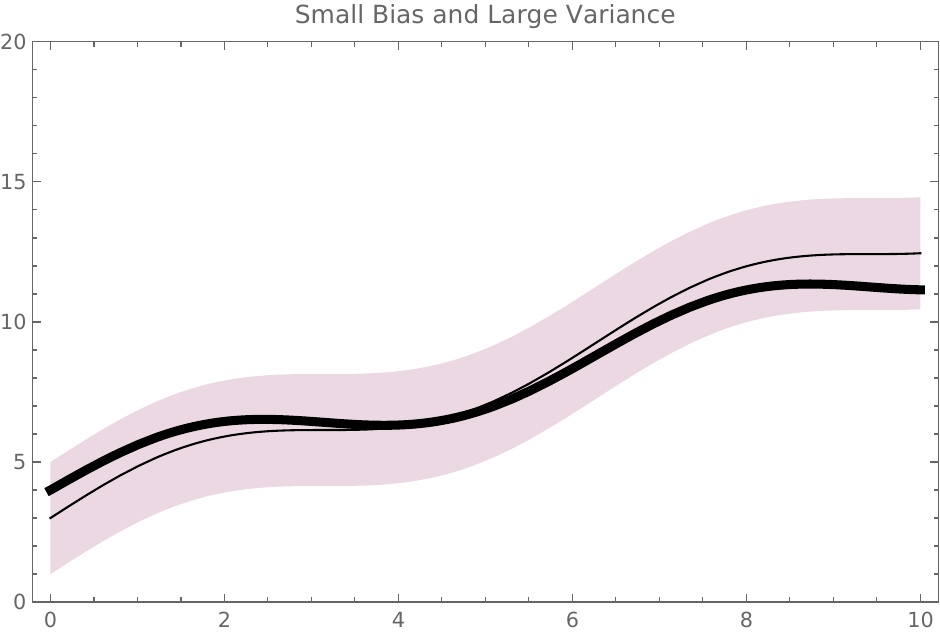}\includegraphics[width = 0.5\textwidth]{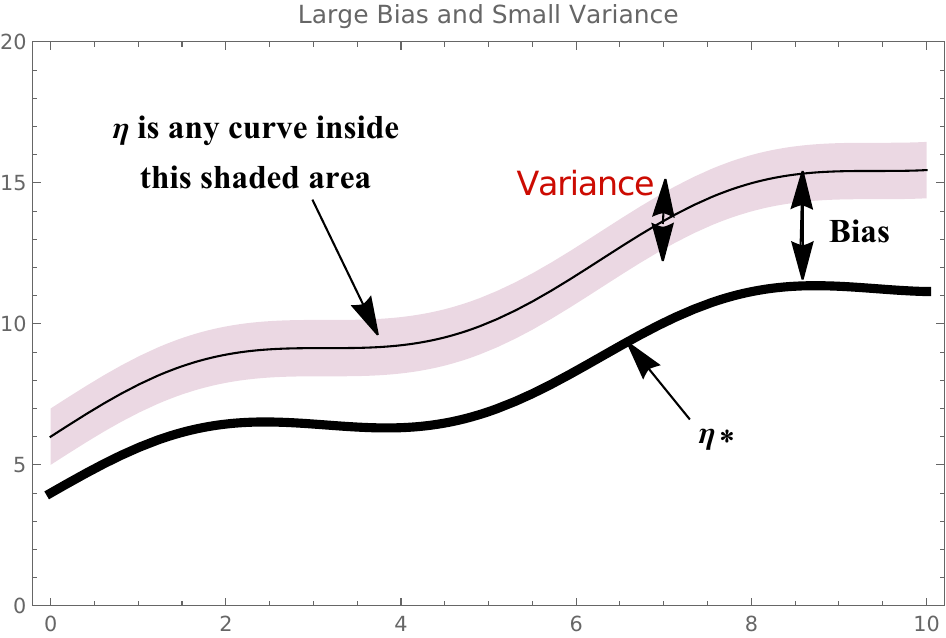}
  \caption{Two regression functions with: low bias and high variance (left); high bias and low variance
    (right)}\label{fig:two-regr-funct}
\end{figure}
Over-training (or overfitting) a particular algorithm is an expression used to describe the complexity of this algorithm, and hence its
capacity, to fit the current training (e.g. getting a very small value of the RSS). Although this seems a success, it is not! This is
because what is required is to have the best performance on the unseen data (the population of testers, as opposed to the training
dataset itself). As explained in Sec.~\ref{sec:motiv-appl}, when a ML algorithm trains on a training dataset there are two things to
realize: (1) the training dataset is taken as an example of the population, and (2) the objective function, to be minimized on this
training dataset, is an estimator of the performance measure that we hope to be minimized on the population. For example, recall that
we used the RSS as an objective function to estimate the MSE as a performance measure.

Overfitting an algorithm results in decreasing the bias of the performance measure and increasing its variance, and vice versa; and
there is always a trade-off between this bias and variance. Before delving into any mathematics, ~\figurename~\ref{fig:two-regr-funct}
qualitatively illustrates this phenomenon for two different ML algorithms (left and right). The bold function $\eta^*$, in both
subfigures, is the conditional expectation, which is the best regression function. Training the algorithm on a training dataset
$\mathbf{tr}$ produces a function $\eta_{\mathbf{tr}}$ that may exist anywhere in the shaded region in the figure. The pointwise mean
of these functions is plotted in light black. The algorithm of the left subfigure produces a mean model
$\MEAN_{\mathbf{tr}}\eta_{\mathbf{tr}}(x)$ that is very close to $\eta^*$ (low bias); however, a single fitted model may exist anywhere
in the wide shaded region (high variance). The algorithm of the right subfigure behaves conversely.

\bigskip

This can be best understood if the KNN is taken as an example. At some point $x_{i}$, the prediction is $\Sigma_{j\in
  N_{K}(x_{i})}y_{j}/K$. The expectation of this regression function is $\Sigma_{j\in N_{K}(x_{i})}\MEAN\left[ {y_{j}}\right] /K$,
while the variance will be $\sigma^{2}/K$ (where the response is assumed to have constant variance $\sigma^{2}$ with the predictor). If
the window size of this rule is squeezed to produce a more complex rule, i.e., $K$ is decreased, the variance will increase, but the
bias will decrease since $\Sigma_{j\in N_{K}(x_{i} )}\MEAN\left[ {y_{j}}\right] /K$ tends to approach $\MEAN\left[ {y_{i}}\right] $. On
the contrary, increasing $K$ obviously decreases the variance, while incorporating many data points whose expectations will be very
likely to vary from $\MEAN\left[ {y_{i}}\right] $, hence the bias increases. This example of KNN is provided
in~\cite{Hastie2009ElemStat}. For more elaboration,~\cite[Ch. 3]{Hastie1990Generalized} review a measure of the complexity of smoothing
functions in terms of an effective number of degrees of freedom.

\bigskip

\begin{figure}[t]\centering
  \includegraphics[width = 0.75\textwidth]{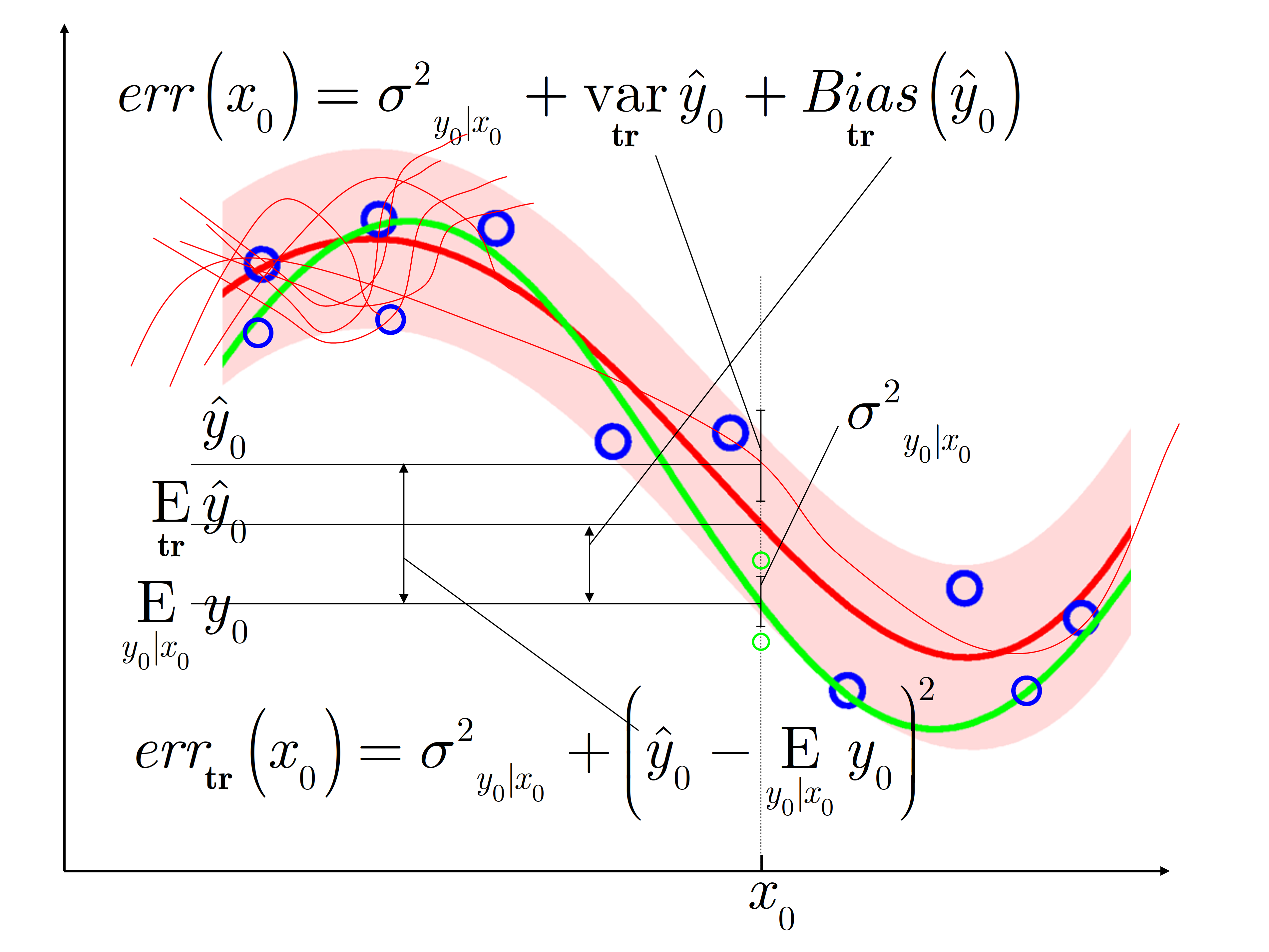}
  \caption{Bias-variance decomposition. A visual illustration using the same colors of the mathematical quantities in
    Eq.~\eqref{eq:VarBiasDecomp}. (Part of the background of the figure, namely, the bold green curve, the bold red curve, and the shaded
    area, is as appears in~\cite[Figure 1.17, pp. 32]{Bishop2006PatRecMachInt}).}\label{fig7}

  \begin{subequations}\label{eq:VarBiasDecomp}
    \begin{align}
      \overline{\Err} &   =\frac{1}{n_{\mathbf{tr}}}\sum_{i\in \mathbf{tr}}\left({\color{red}\widehat{y}_{i}}-{\color{blue}y_{i}}\right)^{2},\\
      \Err_{\mathbf{tr}} & = \MEAN_{x_0,y_0}\left[\left({\color{red}\widehat{y}_0}-{\color{green}y_0}\right)^{2} \right],\\
                     &=  \MEAN_{x_0}\left[ \MEAN_{y_0|x_0}\left({\color{red}\widehat{y}_0}-{\color{green}y_0}\right)^{2} \right]\qquad(\cong \frac{1}{n_{\mathbf{ts}}}\sum_{i\in\mathbf{ts}}\left({\color{red}\widehat{y}_{i}}-{\color{green}y_i}\right) ^{2}),\\
                     &  =\MEAN_{x_0}\left[ \sigma_{Y|X=x_0}^{2}+\left({\color{red}\widehat{y}_0}-{\color{green}\underset{\bm{y_0|x_0}}{\mathbf{E}}{\bm{y_0}}}\right)  ^{2} \right],\\
      \Err &  =\MEAN_{\mathbf{tr}}\Err_{\mathbf{tr}}\qquad (\cong {\frac{1}{M} \sum_{m=1}^{M}\Err_{\mathbf{tr}_{m}}}),\\
                      & = \MEAN_{x_0}\left[ \underset{\text{response variance}}{\sigma_{Y|X=x_0}^{2}}+\underset{\text{Variance}}{\MEAN_{\mathbf{tr}}\left({\color{red}\widehat{y}_0}-{\color{red}\underset{\mathbf{tr}}{\mathbf{E}}{\bm{\widehat{y}_0}}}\right)  ^{2}} + \underset{\text{Bias squared}}{\left({\color{red}\underset{\mathbf{tr}}{\mathbf{E}}{\bm{\widehat{y}_0}}}-{\color{green}\underset{\bm{y_0|x_0}}{\mathbf{E}}{\bm{y_0}}}\right)  ^{2}} \right].\label{eq:VarBiasDecomp-err}
    \end{align}%
  \end{subequations}
\end{figure}
The bias-variance decomposition for a regression function is analyzed quantitatively in Eq.~\eqref{eq:VarBiasDecomp}, and is
illustrated in~\figurename~\ref{fig7}. This figure is conceptually similar to~\figurename~\ref{fig:two-regr-funct}, but with indicating
each component of Eq.~\eqref{eq:VarBiasDecomp} on the figure. For better illustration and pedagogy, the colors of the figure match the
colors of the corresponding terms of the equation as follows: the light red for a trained model, the bold red for its mean over the
population of training datasets, the blue circles for the training dataset (observed response), the green circles for the testing
dataset (observed response), the bold green for the response conditional mean (the best regression function). The symbol $\cong$ is
used to indicate how the corresponding quantity can be estimated from a dataset, regardless whether this is a good estimator or not as
will be explained in the next chapter. The symbol $M$ denotes the number of training datasets
drawn through Monte Carlo (MC) trials.

\bigskip

We end this discussion with the following two questions that pave the road for the subfield of \emph{ML assessment}, the topic of the
next chapter. These questions are valid for any other performance measure, and the error rate is given just as a clear example. The
subfield of \emph{ML assessment} explains different methods for estimating the performance of a ML algorithm, from a given dataset
(because the whole population of testers is unknown) to select among a variety of competing models and assess them, which answers the
following two questions.

(1) How can we minimize the mean error $\MEAN_{\mathbf{tr}}\Err_{\mathbf{tr}}$~\eqref{eq:VarBiasDecomp-err}, where the expectation is
taken over the population of training datasets? As appears from the equation, this error is decomposed to three terms: the response
variance, the model variance, and the model squared bias, respectively. The response variance is the natural variance in the physical
phenomenon that generated the data and is model independent; hence, it is irreducible. Therefore, to minimize the mean error
$\MEAN_{\mathbf{tr}}\Err_{\mathbf{tr}}$, the model complexity should be tuned so that the summation of the bias squared and variance is
minimized. Tersely speaking: too simple (complex) models produce high bias (variance) and low variance (bias); and since the variance
(bias squared) cannot drop below zero, the summation of these two quantities will be high.

(2) Should we design the ML model to minimize the conditional error $\Err_{\mathbf{tr}}$ (conditional on a particular training dataset)
or the mean error $\MEAN_{\mathbf{tr}}\Err_{\mathbf{tr}}$ that involves the bias and variance components?

\subsection{Curse of Dimensionality}\label{subsec:curse}
This expression refers to what may happen when the predictor has high dimensions, i.e., $p$ is too large. The word ``large'' should be
understood relatively to the size $n$ of the available dataset. For illustration, consider smoothing in high dimensions. It will almost
fail because for a fixed number of available observations (the training dataset), the volume size needed to cover a particular
percentage of the total number of observations increases by a power law, and thus exponentially, with dimensionality. This makes it
prohibitive to include the same sufficient number of observations within a small neighborhood, or bandwidth, to smooth the response.
More quantitatively, consider a unit hyper-cube in the $p$-dimensional space containing uniformly distributed observations; the
percentage of the points located inside a hyper-cube with side length $l$ is $l^{p}$. This means, if the suitable bandwidth for a
certain smoother is $l$, the effective number of observations in the $p$-dimensional problem will go as the power $1/p$. This
deteriorates the performance dramatically for $3 < p$. This is why, e.g., the additive model
(Sec.~\ref{subsubsec:additive}) and its variants are expressed as summation of functions of just one dimension. This single
dimension may be just a component of the predictor or a linear combination.

Many other problems occur when $n << p$, including increasing the model variance. In addition, the performance of the model fitted from
a given dataset will not generalize on the population or a future dataset. All of these problems, and others, are indeed related and
connected mathematically to each other, which is out of the scope of the present chapter.

\bigskip

Therefore, a very crucial topic in ML is dimensionality reduction (it is called feature selection in some other communities).
Qualitatively speaking, this means selecting those predictor components that best summarize the relationship between the response and
predictor. In real-life problems, some features are statistically dependent on others; this is referred to as multi-collinearity. On
the other hand, there may also be some components that are statistically independent from the response. These add no additional
information to the problem; thus they serve only as a source of noise.

Several existing approaches aim to reduce the dimensions of the problem. A dimensionality reduction method of course can be considered
as part of the ML algorithm. Therefore, for a given problem, selecting among different methods account as selecting among different
algorithms which is the main topic of the next chapter, as explained at the end of Sec.~\ref{subsubsec:variance}.

\subsection{Performance of Unsupervised Learning}\label{subsec:unsupervised}
It should be noticed that the formal definition of the learning process, discussed thus far in the present chapter, assumed the
existence of a training dataset, $\mathbf{tr}:\left\{ t_{i}=\left( {x_{i},y_{i}}\right),\ i=1,\ldots ,n\right\} $. Each element
$t_{i}$, or sample case, in this set has an already known value for the response variable. This is what enables the learning process to
develop the relationship between the predictor and the response. This is what is called supervised learning. On the contrary, in some
applications the available dataset is described by $\mathbf{tr} :\left\{t_{i}=x_{i},\ i=1,\ldots ,n\right\} $, without any additional
information. This situation is called unsupervised learning. It is usually required in such a situation to understand the structure
of the data from the available empirical probability distribution of the points $x_{i}$. For the special case, where the data come from
different classes, the data will be represented in the hyper $p$-dimensional space, to some extent, as disjoint clouds of data. The
task in this case is called clustering, i.e., trying to identify those classes that best describe, in some sense, the current available
data. More formally, if the available dataset is $\mathbf{X}$, it is required to find the class vector
$\Omega=[\omega_{1},\ldots,\omega_{k}]^{\prime}$ and the clustering function $\eta_{\mathbf{tr}(X)}$, such that a criterion (an
objective function) $\mathcal{J}(\mathbf{X},\Omega)$ is minimized:%
\begin{equation}
  \Omega=\arg\min\left[\mathcal{J}(\mathbf{X},\Omega)\right].\label{eq43}%
\end{equation}
Different criteria give rise to different clustering algorithms. More discussion on unsupervised learning and clustering can be found
in~\cite{Fukunaga1990Introduction, Duda2001PatternClassification, Hastie2009ElemStat}.

It is important to emphasize that although the construction of the supervised and unsupervised rules is quite distinct, the assessment
procedure and the performance measures, including error rate, risk, ROC, AUC, etc., are essentially the same. This is obvious because
the unsupervised rule $\Omega_{\mathbf{tr}}(X)$, regardless of its construction, ultimately provides the same mapping
$\eta_{\mathbf{tr}}(X) \mapsto \{\omega_1,\ldots ,\omega_K\}$ as the supervised rule, which is assigning a class label to a predictor.

\subsection{Classifier Calibration}\label{sec:class-calibr}
As detailed throughout the chapter, the final classifier decision $\eta_{\mathbf{tr}}(x)$ of a classifier is obtained by comparing its
output score $h_ {\mathbf{tr}}(x)$ to a threshold $th$. However, there are two important issues to consider. (1) The scores do not
necessarily equal to the posterior probabilities $\Pr[\omega_i|x]$, which are much more informative than a mere numerical score;
indeed, many classifiers provide score values outside the period $[0,1]$. (2) Scores of two different classifiers cannot be compared,
simply because they are not on the same scale. Classifier calibration is a remedy to these two issues, not naively by linear scaling,
but by providing a one-to-one nonlinear monotonic transformation that maps the output scores to the posterior probabilities. It is
important to observe that this transformation will not affect the performance of the classifier on the population or on a finite
testing dataset. For a formal proof of this result, and for a full account of the calibration process including a recent comparative
study among different calibrators see~\cite{Yousef2021ClassifierCalibration-arxiv}.


  \section{Discussion and Conclusion}\label{sec:concl-advice-pract}
This chapter is intended to provide a pilot view of the field of ML to illustrate how mathematics and intuition together work, which
helps cyberphysical security practitioners, who apply ML in many applications, understand subtle concepts and connect scattered pieces.
The importance of the theoretical aspects of ML are stressed, and demonstrating examples are provided. The mathematical foundations of
the field, along with different methods and construction, have been motivated. Important and fundamental references have been cited for
readers, who are interested in more elaboration.

\bigskip

When it comes to real-life applications, many practitioners leverage some ML approaches, or models, without having the fundamental
rigour or the enough insight, a matter that results in a lot of fallacies and pitfalls. A simple example is the use of complex models,
that have high capacity, relative to the training dataset size. A second example is to perform data preprocessing or transformation
without including the step into the resampling mechanism that estimates the final performance. A third example is thinking of a
particular model or approach as ``magical'' that can consistently outperform others ubiquitously.

``No overall winner'' is a statement that has been touched upon throughout previous sections. If there is no prior information for the
joint distribution between the response and the predictor, and if there is no prior information about the phenomenon to which that
regression or classification will be applied, there is no overall winner among regression or classification techniques. If one method
is found to outperform others in some applications, this is likely to be limited to that very situation or that specific kind of
problem; it may be beaten by other methods for other situations. In the engineering and computer science communities, this concept is
referred to as the \textit{no-free-lunch} theorem~\citep[see][Sec. 9.2]{Duda2001PatternClassification}. This situation holds because
each method makes different assumptions about the application or the process being modeled, and not all real-life applications are the
same. If one or more of the assumptions are not satisfied in a given application, the performance will not be optimal in that setting.
The only unique overall winner is the conditional expectation (for regression) or the Bayes' classifier (for classification) when the
probability distributions are known.

\bigskip

To recap, practitioners are always advised to have a basic level of mathematical rigor and understanding of these foundations, even if
they do not produce research or contribute to theoretical discoveries in the field.






  \section{Acknowledgment}\label{sec:acknowledgment}
The author is grateful to the U.S. Food and Drug Administration (FDA) for funding a very early stage of this chapter, and to Dr. Kyle
Myers for her support. In his memorial, special thanks and gratitude to Dr. Robert F. Wagner, the supervisor and the teacher, or Bob
Wagner, the big brother and friend. He reviewed a very early version of this chapter before he passed away.


  \putbib[booksIhave,publications]
\end{bibunit}

\end{document}